%% file: neurips_2026.tex
\documentclass{article}

\usepackage[eandd, final]{neurips_2026}

\usepackage[most]{tcolorbox} 

\usepackage[utf8]{inputenc} %
\usepackage[T1]{fontenc}    %
\usepackage{url}            %
\usepackage{booktabs}       %
\usepackage{amsfonts}       %
\usepackage{nicefrac}       %
\usepackage{microtype}      %
\usepackage{xcolor}         %
\usepackage{enumitem}
\usepackage{natbib}
\usepackage{marvosym}
\usepackage{graphicx} \usepackage{float}
\usepackage{xcolor}
\usepackage{fontawesome5}
\usepackage{tabularx}
\usepackage{listings}
\usepackage{verbatim}
\usepackage{subcaption}
\usepackage{multirow}
\usepackage{xltabular}
\usepackage{hyperref}

\definecolor{darkred}{rgb}{0.8, 0.1, 0.05}

\newcolumntype{L}{>{\raggedright\arraybackslash}p{2.3cm}}
\newcolumntype{X}{>{\raggedright\arraybackslash}X}

\title{\textsc{AI GameStore}: Scalable, Open-Ended Evaluation of Machine General Intelligence with Human Games}

\author{%
  {\small\textbf{Lance Ying$^{1,2}$, \quad Ryan Truong$^{2}$, \quad Prafull Sharma$^{1}$, \quad Kaiya Ivy Zhao$^{1}$,}} \\
    {\small\textbf{Nathan Cloos$^{1}$, \quad Kelsey R. Allen$^{3}$, \quad Thomas L. Griffiths$^{4}$,}} \\
    {\small\textbf{Katherine M. Collins$^{1,4,5}$, \quad Jos\'e Hern\'andez-Orallo$^{5,6}$, \quad Phillip Isola$^{1}$,}} \\
    {\small\textbf{Samuel J. Gershman$^{2}$, \quad Joshua B. Tenenbaum$^{1}$}} \\[0.75em]
    \small $^{1}$MIT \quad $^{2}$Harvard University \quad $^{3}$University of British Columbia \quad $^{4}$Princeton University \\
  \small $^{5}$University of Cambridge \quad
  $^{6}$Universitat Politècnica de València \\[0.75em]
  \small \faGlobe:\, \href{https://aigamestore.org}{\textcolor{blue}{\texttt{https://aigamestore.org}}} \\[0.4em]
  \small \faGithub:\, \href{https://github.com/lance-ying/aigamestore}{\textcolor{blue}{\texttt{https://github.com/lance-ying/aigamestore}}}
}

\newcounter{savedtable}
\begin{document}

\maketitle

\begin{abstract}
Rigorously evaluating machine intelligence against the broad spectrum of human general intelligence has become increasingly important and challenging in this era of rapid technological advance. Conventional AI benchmarks typically assess only narrow capabilities in a limited range of human activity. Most are also static, quickly saturating as model capabilities improve.  We propose that a more promising way to evaluate human-like general intelligence in AI systems is through a particularly strong form of general game playing: studying how and how well they play and learn to play \textbf{all conceivable human games}, in comparison to human players with the same level of experience, time, or other resources. We define a ``human game'' to be a game designed by humans for humans,
and argue for the evaluative suitability of this space of all such games people can imagine and enjoy --- the ``Multiverse of Human Games''. Taking a first step towards this vision, we introduce the \textbf{\textsc{AI GameStore}}, a scalable and open-ended platform that uses LLMs with humans-in-the-loop to synthesize new representative human games, by automatically sourcing and adapting standardized and containerized variants of game environments from popular human digital gaming platforms. We generated 100 such games based on the top charts of Apple App Store and Steam, and evaluated seven frontier vision-language models (VLMs) on short episodes of play. The best models achieved less than 10\% of the human average score on the majority of the games, and especially struggled with games that challenge world-model learning, memory and planning.  We conclude with a set of next steps for building out the \textbf{\textsc{AI GameStore}} as a practical way to measure and drive progress toward human-like general intelligence in machines.
\end{abstract}

\section{Introduction}\label{sec:intro}

Human intelligence has evolved to solve problems in an open world --- often problems we have never encountered before. The quest to understand and replicate it in machine form has captivated scientists for centuries \citep{turing1950computing, newell1972human,chater2018mind}, and recent AI advances have sparked heated debate about how close machines are to the ``cognitive versatility and proficiency of a well-educated adult'' \citep{hendrycks2025definition,bubeck2023sparks,mitchell2024debates,chen2026does}.

Evaluating whether an AI system has achieved such generality is profoundly difficult. Traditional benchmarks assess performance in isolated domains, such as strategic games \citep{Campbell2002Deep, Silver2016Mastering}, language understanding \citep{wang2018glue}, domain-specific questions \citep{phan2025humanity}, math \citep{Cobbe2021Training}, or code \citep{jimenez2023swe}, and thus only capture fragments of the vast landscape of human capabilities. Even large benchmark collections \citep{srivastava2023beyond} cover a tiny subset of tasks humans can solve, and performance on them remains a poor indicator of open-ended competence \citep{xing2024understanding,li2025gvgai, collins2026expert}. On the other hand, building one benchmark that encompasses all real-world human activities is simply impractical.

We propose that a promising way to assess human-like general intelligence is by studying how well AI systems play and learn to play \textbf{all conceivable human games}---games designed by humans for humans---and comparing their gameplay to human players given the same budget. We call this space the ``Multiverse of Human Games.'' Games are cultural artifacts designed as miniatures of real-world activities, covering nearly every human skill from strategic planning to social reasoning to navigating complex physical environments. To excel across all conceivable human games is to possess a diverse set of cognitive capabilities needed for flourishing in the human world.

While appealing in principle, working with actual human games poses significant challenges, including interface heterogeneity, intellectual property constraints, and data contamination risk. In this work, we take a first concrete step by proposing the \textbf{\textsc{AI GameStore}}, a platform that leverages LLMs with humans-in-the-loop to source and adapt diverse human games from popular digital marketplaces into standardized evaluation environments, thus producing a never-ending game benchmark for AI models. As a proof of concept, we curated 100 games and compared seven frontier vision-language models (VLMs) against 106 human participants. We find that state-of-the-art models achieved less than 10\% of the human average score on the majority of games, while taking 15-20x longer to complete. They especially struggled with games challenging world-model learning, memory, and planning.

In summary, we make the following contributions: (1) We introduce the Multiverse of Human Games as a route to measuring human-like general intelligence; (2) We propose the \textbf{\textsc{AI GameStore}}, a scalable, open-ended evaluation platform sampling from this space; (3) We generate 100 games adapted from Apple App Store and Steam top charts; (4) We evaluate frontier VLMs against human play, revealing significant cognitive capability gaps.

\section{Measuring general intelligence with the Multiverse of Human Games}\label{sec:multiverse_games}

A central argument of this paper is that the space of all conceivable human games provides a uniquely comprehensive set of tasks for evaluating machine general intelligence. By human games, we mean games that humans intentionally have designed for themselves or other humans to play. These games are by definition enjoyable and learnable by (at least some) people. The space of
all conceivable human games is infinite and open-ended: these are all the games that humans could create and that other humans could enjoy. By the ``Multiverse of Human Games'', we mean not only this space but also the associated distribution of how likely they are to be created and spread by humans. This strongly deviates from universal distributions of tasks where most of the mass of the distribution would be composed of games humans would never build or enjoy playing \citep{legg2007universal}.  We propose that a promising way to evaluate how well machines are approaching human-level general intelligence is by testing how well a machine can learn to play representative samples of the Multiverse of Human Games relative to a typical human player when given the same gameplay budget.

This paradigm builds on a long and rich tradition of using games and general game playing to study intelligence \citep{cleveland1907psychology, genesereth2005general,schaul2011measuring, OpenAIUniverse2016,hernandez2017new, perez2019general,hafner2021benchmarking, allen2024using, paglieri2024balrog, collins2025people, ying2025assessing, li2025gvgai,wang2025game,lehrach2025code,magne2026nitrogen, google2026gamearena, arcagi3_2026} (See related work in section \ref{sec:related_work_main}.) Our aim here is to extend these efforts with a formulation and technical approach that scales to the full space and distribution of all games designed or to be designed by humans -- although in this paper we can implement only a first, very small step towards this goal. In the following sections, we will discuss why the Multiverse of Human Games is a good way to evaluate truly general intelligence as well as the practical challenges and strategies for operationalizing this evaluation paradigm with the \textbf{\textsc{AI GameStore}}. 

\subsection{Why is this a good measurement of general intelligence?}

The proposition that all conceivable human games serve as a robust proxy for human-like general intelligence is rooted in the teleology of play: humans design and propagate games to prepare themselves for the multifaceted challenges of dynamic environments. Play is a foundational mechanism for learning across species \citep{chu2020play, smith1982does, burghardt2024animal}, and gameplay is associated with improvements in executive functions, spatial reasoning, and attentional control \citep{spence2010video, feng2007playing}.

Furthermore, games serve as vital cultural artifacts \citep{chu2024praise}. They are not merely pastimes but are sophisticated vehicles for cultural transmission. By abstracting and containerizing real-world complexities --- ranging from the strategic planning in multi-party conflicts to the nuanced social dynamics of RPG games --- humanity has collectively created a curriculum for training and preparing individuals for surviving and adapting in the open world. Games often pass down through generations and spread across cultures: invented thousands of years ago, Go and Chess are still enjoyed by millions of players today; the Olympic Games, which first emerged from the cradle of ancient Greek civilization, have evolved into a grand cross-cultural human enterprise with events practiced and watched by people around the world. Consequently, the space of all conceivable human games represents a distilled, concentrated library of the essential skills required to navigate the world that humans live in. To excel in this space of games is, by design, to exhibit the core tenets of human-like general intelligence.

While the Multiverse of Human Games is unbounded, all games on digital gaming platforms provide a finite set of well-defined instances covering diverse cognitive skills. However, directly using these millions of games as a benchmark faces major roadblocks: (1) \textbf{copyright restrictions} preventing use in public benchmarks; (2) \textbf{platform heterogeneity} across engines, OSes, and APIs; (3) \textbf{privacy constraints} on human gameplay data; (4) \textbf{latency mismatch} between real-time games and slow AI model inference; and (5) \textbf{dataset contamination risk} since training data is undisclosed.

We propose an alternative: the \textbf{\textsc{AI GameStore}}, a meta-benchmark that uses an automated pipeline to source and rebuild synthetic games representing standardized versions of popular human games, suitable for tractable AI evaluation and human comparison. While we show this approach is highly scalable, we acknowledge that it currently covers only a limited region of the full Multiverse, namely casual digital games, and discuss the relationship between these spaces further in Appendix \ref{sec:venn_diagram}.

\begin{figure}
    \centering
    \includegraphics[width=0.9\linewidth]{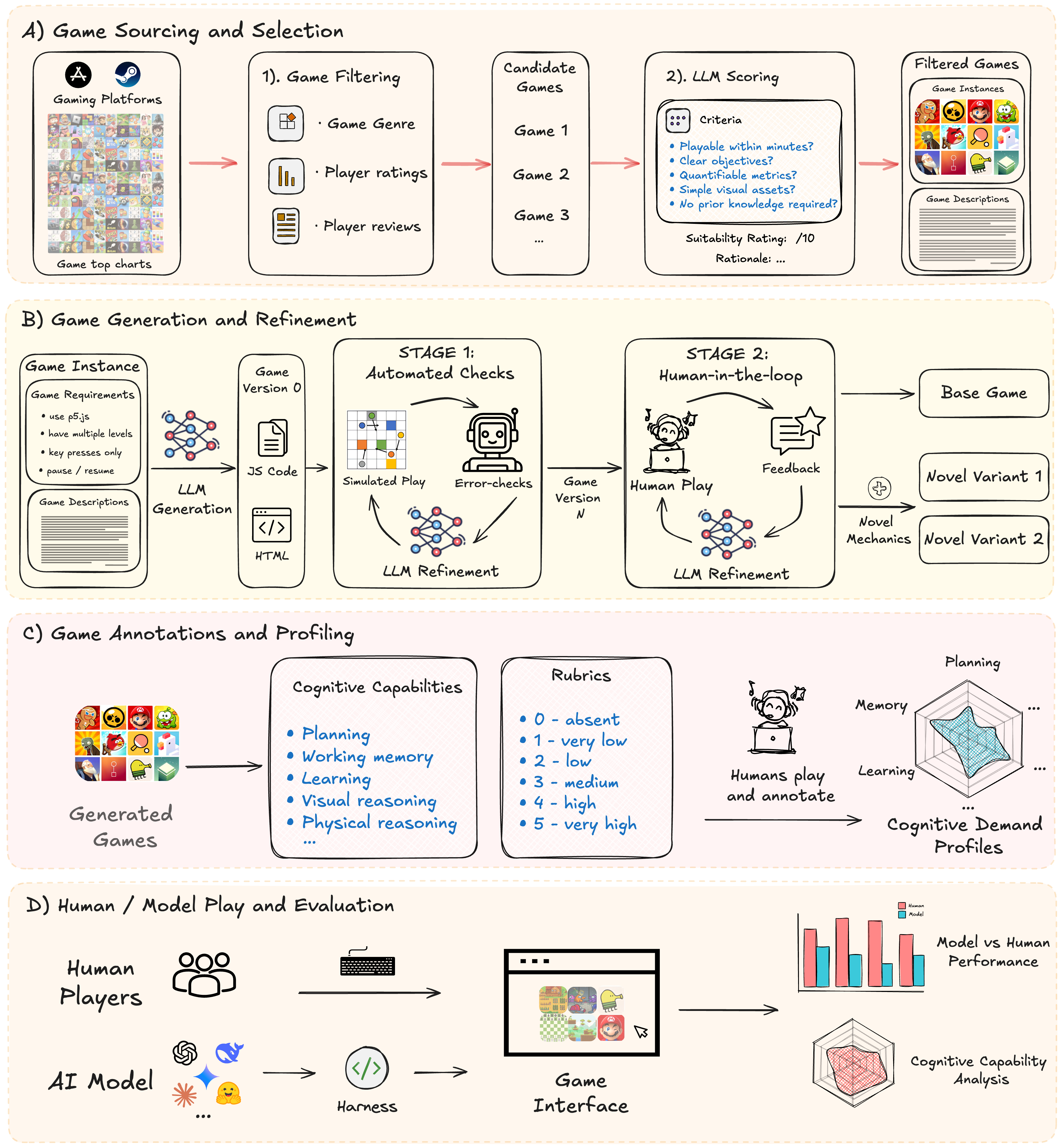}
    \caption{The \textbf{\textsc{AI GameStore}} pipeline: \textbf{a)} Games are sourced from digital marketplaces and filtered by ratings and LLM-scored suitability. \textbf{b)} An LLM generates games from descriptions, refined via automated testing and human-in-the-loop feedback to produce base games and novel variants. \textbf{c)} Games are annotated by humans on cognitive capability rubrics. \textbf{d)} AI models and humans play through a standardized interface for comparative evaluation.}
    \label{fig:generation_pipeline}
\end{figure}

\section{The \textbf{\textsc{AI GameStore}}}\label{sec:ai_gamestore}

\begin{figure}
    \centering
    \includegraphics[width=\linewidth]{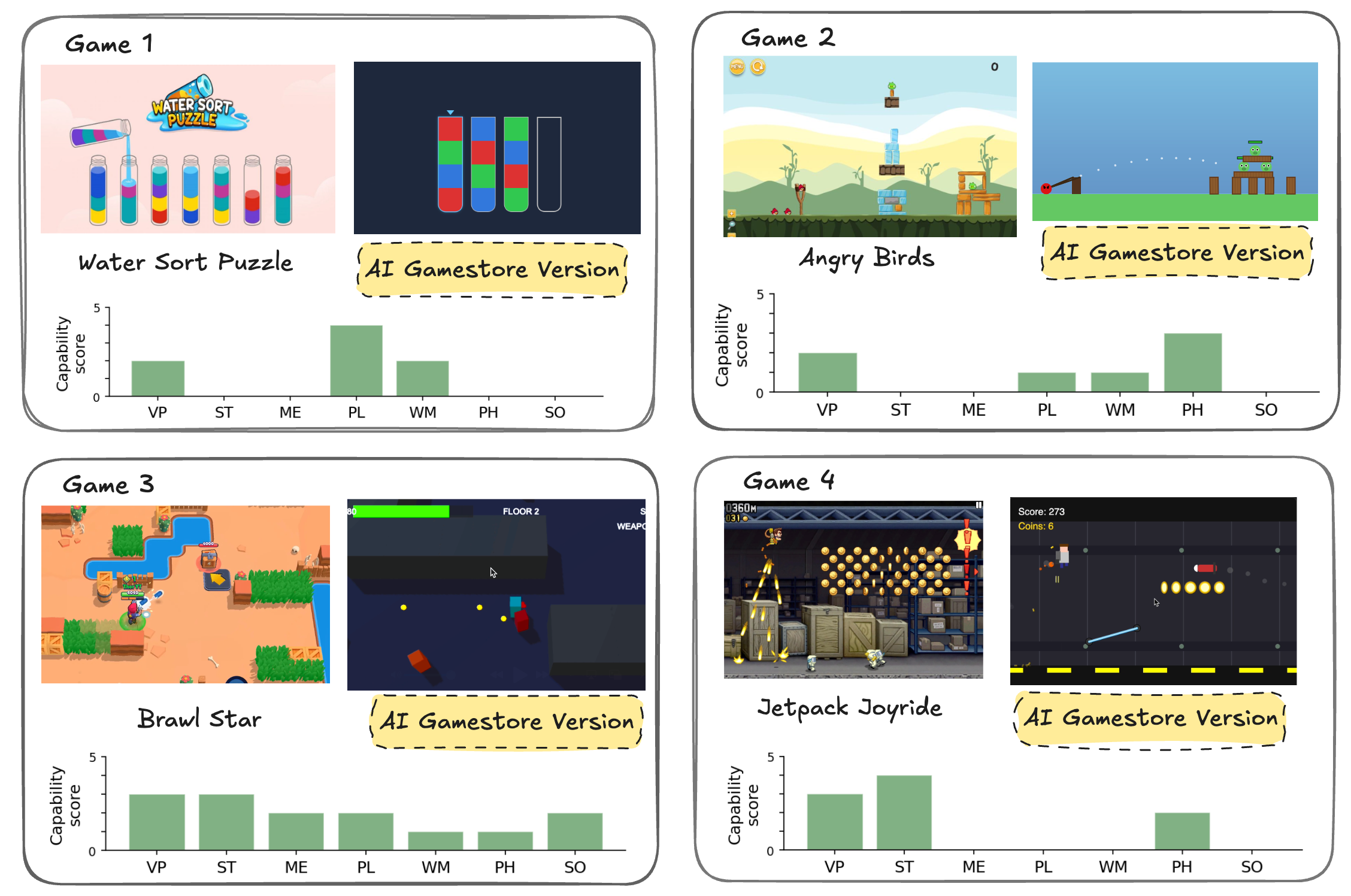}
    \caption{Four example games from the Apple App Store and their \textbf{\textsc{AI GameStore}} adaptations, with cognitive demand annotations. Games and model vs human play videos available at \href{http://aigamestore.org}{aigamestore.org}.(VP = Visual Processing; ST = Spatial-temporal Coordination; ME = Memory; PL = Planning; WM = World Model Learning; PH = Physical Reasoning; SO = Social Reasoning.)}
    \label{fig:example_games}
\end{figure}

The \textbf{AI GameStore} offers a unified sandbox environment designed for evaluating models on standardized adaptations of popular digital games (See Figure \ref{fig:example_games}). Unlike many existing generative game evaluation frameworks (e.g. \citep{cobbe2020leveraging, verma2025measuring}), \textbf{\textsc{AI GameStore}} aims to source and generate human games that are designed and enjoyed by humans.

To do so, the \textbf{\textsc{AI GameStore}} utilizes a scalable, semi-automated pipeline for the procedural generation and refinement of evaluation tasks, as visualized in Figure \ref{fig:generation_pipeline}. This pipeline operates across four distinct stages:

\begin{itemize} 

\item \textbf{Stage 1: Sourcing and suitability filtering} High-quality game candidates are harvested from existing gaming platforms and subjected to a multi-stage filter based on player engagement metrics and an LLM-driven suitability scoring system. 

\item \textbf{Stage 2: Game generation and refinement} Utilizing filtered game descriptions, an LLM generates a functional JavaScript codebase, written in p5.js, matter.js, and/or three.js.  This draft undergoes automated unit testing via simulated play to ensure mechanical stability and basic responsiveness to input. The functional code is subjected to a secondary refinement phase where human participants provide natural language feedback to correct mechanical issues, increase playability and propose novel gameplay variations. 

\item \textbf{Stage 3: Game annotation and profiling} To characterize the latent cognitive demands of the benchmark, each finalized game is subjected to a human annotation process. Expert annotators evaluate the tasks across a multi-dimensional cognitive taxonomy using a 0-5 scale. These profiles allow for the disentanglement of complex model behaviors by mapping performance failures to specific cognitive demands, ensuring that the \textbf{\textsc{AI GameStore}} serves as both a benchmark and a diagnostic tool for understanding machine intelligence.

\item \textbf{Stage 4: Model evaluation} In the final stage, both human players and various AI models interact with the games through a gameplay interface. AI models are integrated via a specialized harness to ensure standardized interaction. The resulting game output is used to compute aggregate human vs model performance as well as deeper analysis on model's cognitive capabilities.
\end{itemize}

This hybrid approach is highly efficient: generating and refining a new game takes approximately 30 minutes on average. The framework creates a living evaluation suite, as new games from marketplaces can be ingested and human participants can create novel variants, ensuring robustness to overfitting and saturation.

\subsection{Sourcing games from digital game marketplaces}\label{sec:sourcing_games}

The \textbf{\textsc{AI GameStore}} draws content from popular digital game marketplaces, specifically focusing on games from the Apple App Store and Steam. We choose these platforms as primary sources because: 1) they are immensely popular with hundreds of millions of monthly active users worldwide; 2) games are created by a large diverse cohort of developers with diverse themes and concepts; 3) new games are constantly being developed and released, which mitigates saturation; and 4) these platforms maintain rigorous review systems that screen out poorly designed games.

We sampled 7,500 games from these platforms with diverse genres and themes (see Appendix \ref{sec:source_games} for details). We filtered the games based on popularity and diversity, retaining those with at least 10,000 reviews and an average rating over 4.5 out of 5. We then used Gemini 2.5 Flash to score each game on suitability for conversion, based on: 1) whether the game can be reasonably played within a few minutes, 2) can be expressed in JavaScript with specified libraries, 3) can have a quantifiable metric for evaluating performance, and 4) does not require extensive game-specific knowledge (e.g.\ poker). After filtering, we retained 100 games for generation.

\subsection{Game adaptation and construction} \label{sec:game_adaptation}

Inspired by recent work on LLM-powered game generation \citep{todd2024gavel, nasir2024word2world, kanervisto2025world}, we prompted \textsc{Claude-sonnet-4.5} to generate a game based on each filtered game description. To facilitate practical and scalable evaluation, we designed a detailed game spec (Appendix \ref{sec:game_specs}) requiring all games to be written in JavaScript with keyboard-only controls, pausability, scoring mechanics, and multiple game levels. This ensures all games can run on a web portal and be scored for cross-model and human comparison in a standardized way.

To ensure games are playable and capture the core mechanics of the original, the pipeline contains two refinement stages. First, an LLM generates test scripts based on the game source code to simulate different actions and detect bugs in the game script. Upon catching bugs, the LLM fixes them until the game passes all tests. Then in the second stage, the game is refined with a human in the loop via a customized interface (Appendix \ref{sec:refine_interface}). The human player plays, evaluates, and gives feedback to the LLM until a good base game is generated. Each refinement step takes about 2 minutes, averaging 4.7 steps across all 100 games. On average, a game can be generated within 30 minutes.

Additionally, human players may generate variants of each game by proposing novel game mechanics using the same interface (Figure \ref{fig:novel_variant}). This allows many valid variations %
per source game, thus increasing the diversity and richness of the game suite and mitigating benchmark saturation.

\subsection{Game annotation and profiling}

We annotated each game across seven cognitive capability dimensions: Visual Processing, Spatial-temporal Coordination, Memory, Planning, World Model Learning, Physical Reasoning, and Social Reasoning. These categories draw from prior literature \citep{zhou2025general, hendrycks2025definition, ying2025assessing} with additions suited to dynamic gameplay. We developed behaviorally anchored five-point rubrics for each capability. Three annotators independently rated every game before deliberating to resolve disagreements. The full taxonomy and rubrics are provided in Appendix \ref{sec:rubrics}; inter-annotator reliability is reported in Appendix \ref{sec:annotation_reliability}. The 100 games show significant diversity, with many requiring multiple sophisticated capabilities (Figure \ref{fig:example_games}).

\subsection{Model evaluation}

The final stage involves evaluating AI models against human players. Both interact with games through a unified interface with the same time budget, enabling direct performance comparison. Models are integrated via a specialized evaluation harness (Section \ref{sec:experiments}). By leveraging our cognitive demand annotations, we can probe why models fail at specific tasks and perform fine-grained analysis of their cognitive strengths and weaknesses.

\subsection{Game release and updates}
To mitigate overfitting and contamination, we release 10 \% of the game corpora to public along with the human participants' gameplay data (available on our \href{https://aigamestore.org}{website}); the remaining constitute a private test set. Statistics comparing public and full sets appear in Appendix \ref{sec:public-games}, showing similar distribution of difficulty and model performance. As new games are developed and released, the pipeline mentioned above will continue sourcing and generating new games as a living benchmark.

\section{Model and human experiments}\label{sec:experiments}

\subsection{Human gameplay experiment}
We recruited 106 participants from Prolific (mean age = 38.81, 58 male, 46 female, 2 non-binary) for a human experiment approved by an IRB. Each participant played 10 distinct, randomly assigned games for 120s each at 60 frames per second. Each participant received a single 120s episode per game and never returned to that game in a later session; any learning or retries therefore occurred only within the episode. Scores were collected every 30 frames with full gameplay recordings (Appendix \ref{sec:protocol_sensitivity}). In addition, we asked participants to rate how fun and how challenging each game was, and, in a separate study, to judge how closely the adapted games matched their original source games, compared with non-source games. We find that the games were rated as moderately fun ($52.4 \pm 13.2$ on a 0--100 scale) and challenging ($59.3 \pm 13.9$; Appendix~\ref{sec:public-games}), and that participants rated source-game clips as far more similar than non-source clips in visuals (79.7 vs.\ 11.5), mechanics (85.6 vs.\ 16.8), and cognitive demands (80.6 vs.\ 29.9; Appendix~\ref{sec:fidelity_study}), indicating that the adapted games are engaging for human players and preserve the visual structure, core mechanics, and cognitive demands of their source games.

\subsection{Model gameplay experiment}

We evaluate seven frontier VLMs: \textsc{GPT-5.2}, \textsc{GPT-5-mini}, \textsc{Gemini-2.5-Pro}, \textsc{Gemini-2.5-Flash}, \textsc{Claude-Opus-4.5}, \textsc{Qwen-3-VL-32B}, and \textsc{Llama-4-Maverick}, each run three times per game with default configurations.

Since models cannot interact in real-time, we built a harness (Figure \ref{harness} in Appendix) inspired by \cite{zhang2025videogamebench}: the game pauses every second, the model provides five lists of actions (one list per 0.2s segment), and actions are applied upon resuming. This continues for 120 API calls (2 minutes of game time). Each prompt includes: game description, the model's scratchpad (serving as memory), recent screenshots and actions, and available actions. The model returns an updated scratchpad, actions, and reasoning (details in Appendix \ref{sec:harness}).

\section{Results}\label{sec:results}

\begin{figure}
    \centering
    \includegraphics[width=0.85\linewidth]{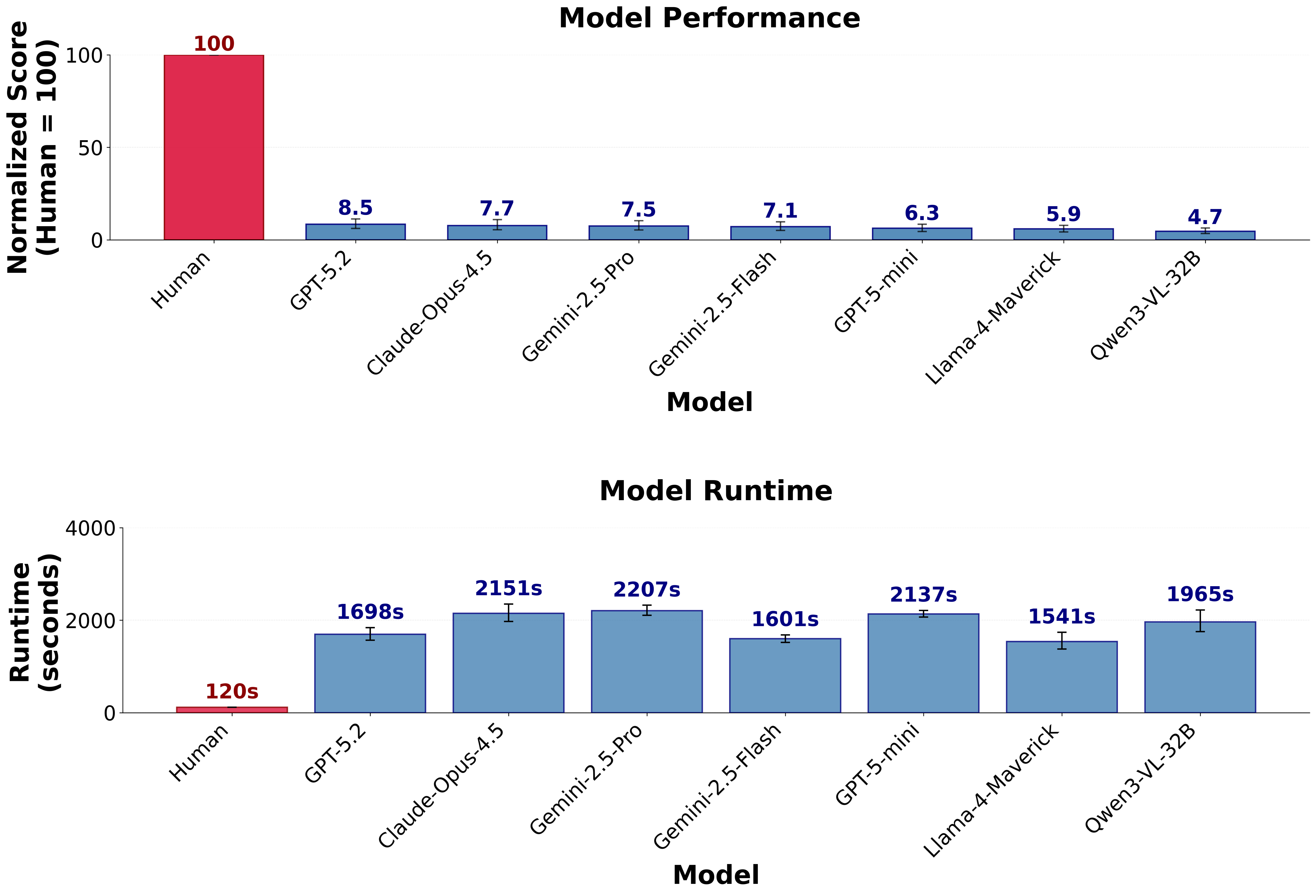}
    \caption{Performance (top) and runtime (bottom) of VLMs vs.\ humans on 100 games. Scores are normalized by human median (= 100) and aggregated via geometric mean. The best model reaches only 8.5\% of human median score. Models require 12-18x longer than humans' play. Error bars indicate 95\% CI.}
    \label{fig:model_performance}
\end{figure}

We report the geometric mean of each models' human-relative performance across all 100 games in Figure \ref{fig:model_performance}. Following prior work comparing AI models with humans on large suites of video games \citep{tsividis2021human}, we chose to report the geometric mean because we are interested in how models perform across many tasks relative to typical human players, and these scores can be heavily skewed. To account for the heterogeneous scoring scales across the game suite, we normalize all model scores by the median human performance in each game, where the human median for each game is set to 100, and we cap model scores between 1 to 10000 before computing geometric means:

\begin{align*}
\text{Normalized Score} &= \text{clip} \left(100 \times \frac{\text{Raw Game Score}}{\text{Human Median Score}}, 1, 10000 \right) .
\end{align*}

While the evaluated models demonstrate the ability to navigate and interact with most game environments, a substantial performance gap remains between AI agents and human participants. State-of-the-art models like \textsc{GPT-5.2}, \textsc{Gemini-2.5-Pro}, and \textsc{Claude-Opus-4.5}, 
all achieve geometric mean scores of less than $10\%$ of the human baseline (see Figure~\ref{fig:model_performance}). The differences between the top 6 models are not statistically significant ($p<0.05$). 

\begin{figure}
    \centering
    \includegraphics[width=\linewidth]{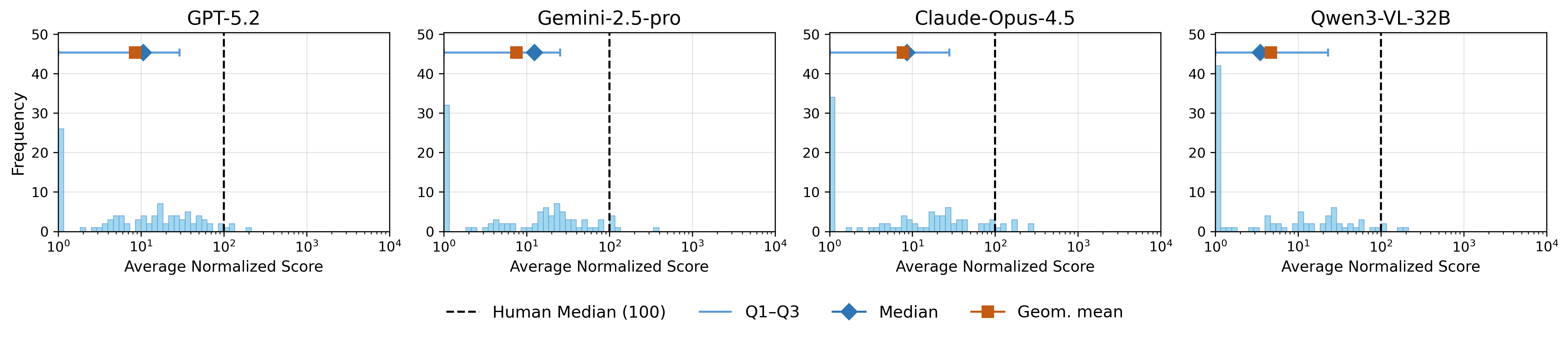}
    \caption{Score distributions across 100 games (normalized, human median = 100, capped at 10,000). All models show a bimodal pattern: near-zero on 30-40\% of games, moderate progress on the rest.}
    \label{fig:score_distribution}
\end{figure}

All models exhibit a distinct bimodal score distribution (Figure \ref{fig:score_distribution}). For roughly two-thirds of games, models make some progress---they can even surpass the median human score on a handful of easy casual games where a simple strategy executed quickly suffices. However, for most games where models make progress, they score significantly worse than median humans (10-30\% of the human level) while taking much longer to reason about each move. Then for the remaining 30-40\% of games, models struggle fundamentally and fail to achieve any meaningful progress at all ($<$1\% of human median).

\begin{figure}
    \centering
    \includegraphics[width=\linewidth]{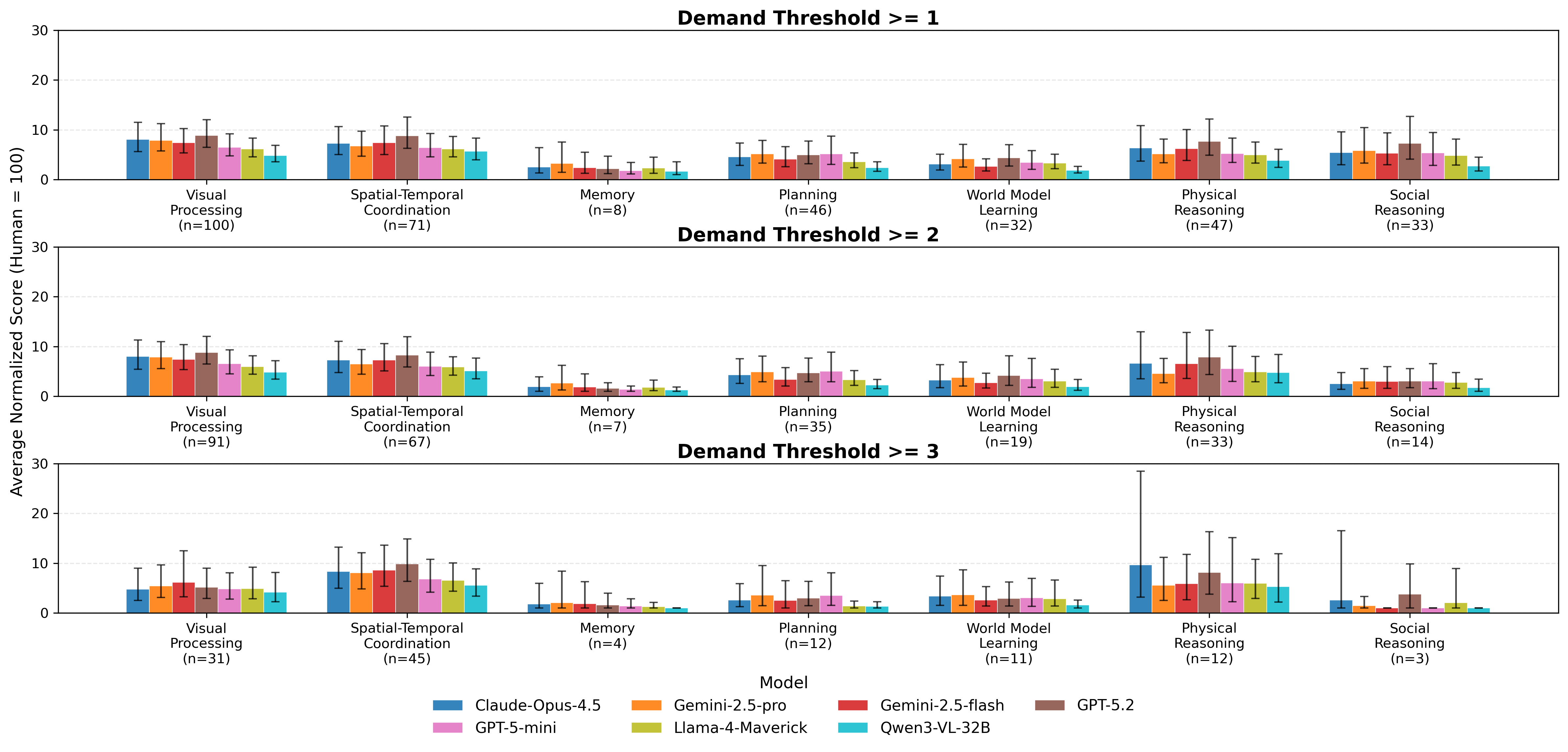}
    \caption{Model performance (geometric mean) grouped by cognitive capability at three demand thresholds. Games are included if their demand for a capability meets or exceeds the threshold. Models struggle most with Memory, Planning, and World Model Learning. Error bars indicate 95\% CI; game counts are shown in brackets.}
    \label{fig:cognitve_performance}
\end{figure}

While these aggregate scores give some indication of the human-AI gap, they provide limited visibility into \emph{what} makes different games challenging for models. To gain more insight, we break down performance by cognitive capability profiles (Figure \ref{fig:cognitve_performance}), which shows the geometric mean of model performance on games grouped by each cognitive capability. Models consistently perform below human baselines across all tested capabilities. Notably, Memory, Planning, and World Model Learning represent the most significant bottlenecks. This suggests that models still struggle with maintaining information across timesteps even with a scratchpad, and with solving problems in ambiguous domains with underspecified rules and mechanics. Additionally, model performance decreases significantly on games that require planning and social reasoning as games demand more sophisticated levels of these capabilities. Example model failure cases are shown in Appendix \ref{sec:failure_case}.

One possible explanation for poor model performance is that the harness does not allow sufficiently fast reactions, since the model is queried for 5 sets of actions every game second. However, model performance on games requiring low spatial-temporal coordination (demand $\leq$ 2), such as puzzle and turn-based strategy games, shows little difference from aggregate performance (Appendix \ref{sec:additional_results}), suggesting the gap is largely driven by cognitive factors rather than solely due to reaction time.

\begin{figure}
    \centering
    \includegraphics[width=\linewidth]{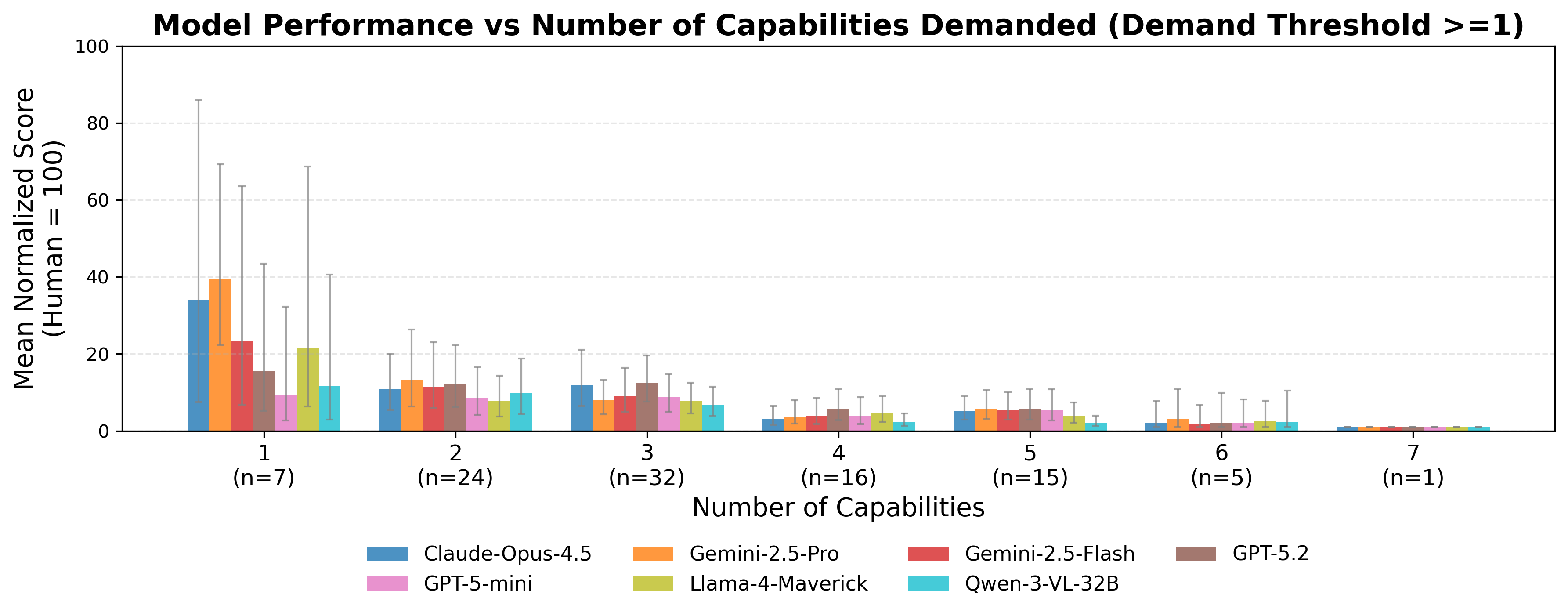}
    \caption{Model performance vs.\ number of cognitive capabilities demanded. Models approach human levels only on games requiring one capability; performance drops sharply as more capabilities must be integrated. Error bars indicate 95\% CI.}
    \label{fig:skill_count}
\end{figure}

We also assessed how model performance varies with the number of capabilities a game demands (Figure \ref{fig:skill_count}). When models perform close to median human levels, it is typically only for games that place a high demand on just one cognitive capability. As games demand more distinct capabilities, performance drops significantly, indicating that success requires not only competence in each capability but the ability to integrate them.

Cumulative score trajectories on our 10 public games are shown in the Appendix (Figure \ref{fig:model_vs_human}), which further illustrate this gap: while humans make steady progress, models typically stagnate early or fail entirely on a significant fraction of games. 

Lastly, we also evaluated the average runtime of all models relative to human players (Figure \ref{fig:model_performance}). Models all take significantly longer than humans to complete each game. This is because the models spend a few minutes thinking, in addition to typically a few seconds of response latency per query; as a result, many models spend at least 20 minutes on the game, while humans play the games within 2 minutes. To achieve genuine human-level general intelligence, future AI models should strive for real-time integrated systems that aim to play games in human-like way, performing perception, thinking, decision-making simultaneously and within a strict time budget.

\section{Discussion and future directions}\label{sec:discussion}

We have argued that evaluating AI systems on the full space of possible human games is a promising way to measure progress toward human-like general intelligence, and proposed the \textbf{\textsc{AI GameStore}} as a first step towards this vision. Even on our current suite of relatively simple games, frontier VLMs achieved less than 10\% of median human scores while taking 15-20x longer, struggling particularly with episodic memory, world-model learning, and long-horizon planning.

In its current state, the \textbf{\textsc{AI GameStore}} is still relatively primitive, and limited in the capacities it can assess and the insights it can yield. Nonetheless, we see this work as evidence %
that it is practical and valuable to build a scalable and open-ended AI evaluation approach based on synthetic versions and variants of actual human games. To advance this approach towards a truly comprehensive framework for evaluating general human-like intelligence (as indicated in Figure \ref{fig:venn_diagram}), our future work must extend the \textbf{\textsc{AI GameStore}} in many directions including the following: (1) \textbf{expanding game diversity}, particularly games requiring long-horizon planning and sophisticated multi-agent social reasoning \citep{leibo2021scalable}; (2) \textbf{generating more complex games} with longer timescales and richer storylines that test sustained memory and world-model formation; (3) \textbf{automated level generation} with better difficulty control \citep{todd2023level, collins2025gamecreation}; and (4) \textbf{deeper capability analysis} using more sophisticated profiling techniques such as measurement layouts \citep{mlayoutstutorial2024} or the ADeLe methodology \citep{zhou2025general} adapted for sequential decision-making.

While our platform is only a modest beginning, we hope it serves as a catalyst for building more general, scalable, open-ended AI evaluations---and a step towards general-purpose agents capable of interacting flexibly, safely, and robustly with human beings in a human world.

\section{Related work}\label{sec:related_work_main}

\paragraph{AI evaluation.} Measuring progress toward general intelligence has traditionally relied on curated benchmarks targeting specific domains---language understanding \citep{wang2018glue}, reasoning \citep{phan2025humanity}, mathematics \citep{Cobbe2021Training}, and code \citep{jimenez2023swe}. Large multi-task suites such as BIG-bench \citep{srivastava2023beyond} enable standardized comparison but remain fundamentally static. As models increasingly optimize for these benchmarks, concerns about saturation, contamination, and narrow capability measurement have grown \citep{collins2026expert, hendrycks2025definition}. Recent work introduces capability-oriented frameworks \citep{mlayoutstutorial2024, zhou2025general} and interactive benchmarks in game-like environments such as ARC-AGI-3~\citep{arcagi3_2026}, Balrog~\citep{paglieri2024balrog}, and AutumnBench~\citep{warrier2025benchmarking}. The \textbf{\textsc{AI GameStore}} extends these work by continually generating new tasks grounded in diverse human-designed games, creating a living meta-benchmark rather than expanding static datasets.

\paragraph{General game playing.} Games have long served as central testbeds for studying intelligence due to their well-defined rules, measurable performance, and diversity of cognitive demands \citep{newell1965example, schaul2011measuring, gobet2004moves}. Early milestones demonstrated superhuman performance in chess \citep{Campbell2002Deep} and Go \citep{Silver2016Mastering} through search and reinforcement learning. Recognizing the limitations of game-specific systems, the General Game Playing paradigm \citep{genesereth2005general} and platforms like GVGAI \citep{perez2019general} evaluate agents across multiple games without game-specific engineering. Recent work extends this to foundation models and LLM-based agents on diverse video games \citep{zhang2025videogamebench, li2025gvgai, wang2025game, lehrach2025code}. We extend this tradition by shifting focus from synthetic environment families to the full distribution of games created and played by humans, constituting a stronger notion of general game playing: the objective is to learn and play the space of human-designed games under human-like constraints.

\paragraph{LLM-guided game generation.} LLMs can translate natural language descriptions into playable environments \citep{nasir2024word2world}, support procedural content generation such as level design \citep{todd2023level}, and generate full games through iterative pipelines \citep{todd2024gavel, kanervisto2025world}. Generated environments have been proposed as evaluation tools \citep{cobbe2020leveraging, verma2025measuring}, but fully automated generation introduces challenges related to playability, representativeness, and evaluation validity---generated tasks may lack meaningful structure or drift from human-relevant activities. The \textbf{\textsc{AI GameStore}} anchors generation in existing popular game concepts from gaming platforms and combines LLM generation with human-in-the-loop refinement, preserving playability, diversity, and evaluative relevance.

\section*{Acknowledgments and disclosures}
This work is funded in part by AFOSR, ONR through the Science of AI, and MURI programs, a Schmidt AI2050 Fellowship to JBT, and the Siegel Family Quest for Intelligence at MIT. Research was also sponsored by the Department of the Air Force Artificial Intelligence Accelerator and was accomplished under Cooperative Agreement Number FA8750-19-2-1000. The views and conclusions contained in this document are those of the authors and should not be interpreted as representing the official policies, either expressed or implied, of the Department of the Air Force or the U.S. Government. The U.S. Government is authorized to reproduce and distribute reprints for Government purposes notwithstanding any copyright notation herein. 

JHO's research is supported by OpenAI's grant to the `AI Progress through the Lens of Predictable AI Ecosystems' programme, which is based at the Leverhulme Centre for the Future of Intelligence at the University of Cambridge. KMC acknowledges support from the NSF SBE SPRF, King's College Cambridge, and the Cambridge Trust.

\textbf{Competing interests.} The authors declare no competing interests.

\bibliography{bibliography}
\bibliographystyle{plain}
\newpage
\appendix

\section{Scope of the AI GameStore}\label{sec:venn_diagram}

\begin{figure}[h!]
    \centering
    \includegraphics[width=0.9\linewidth]{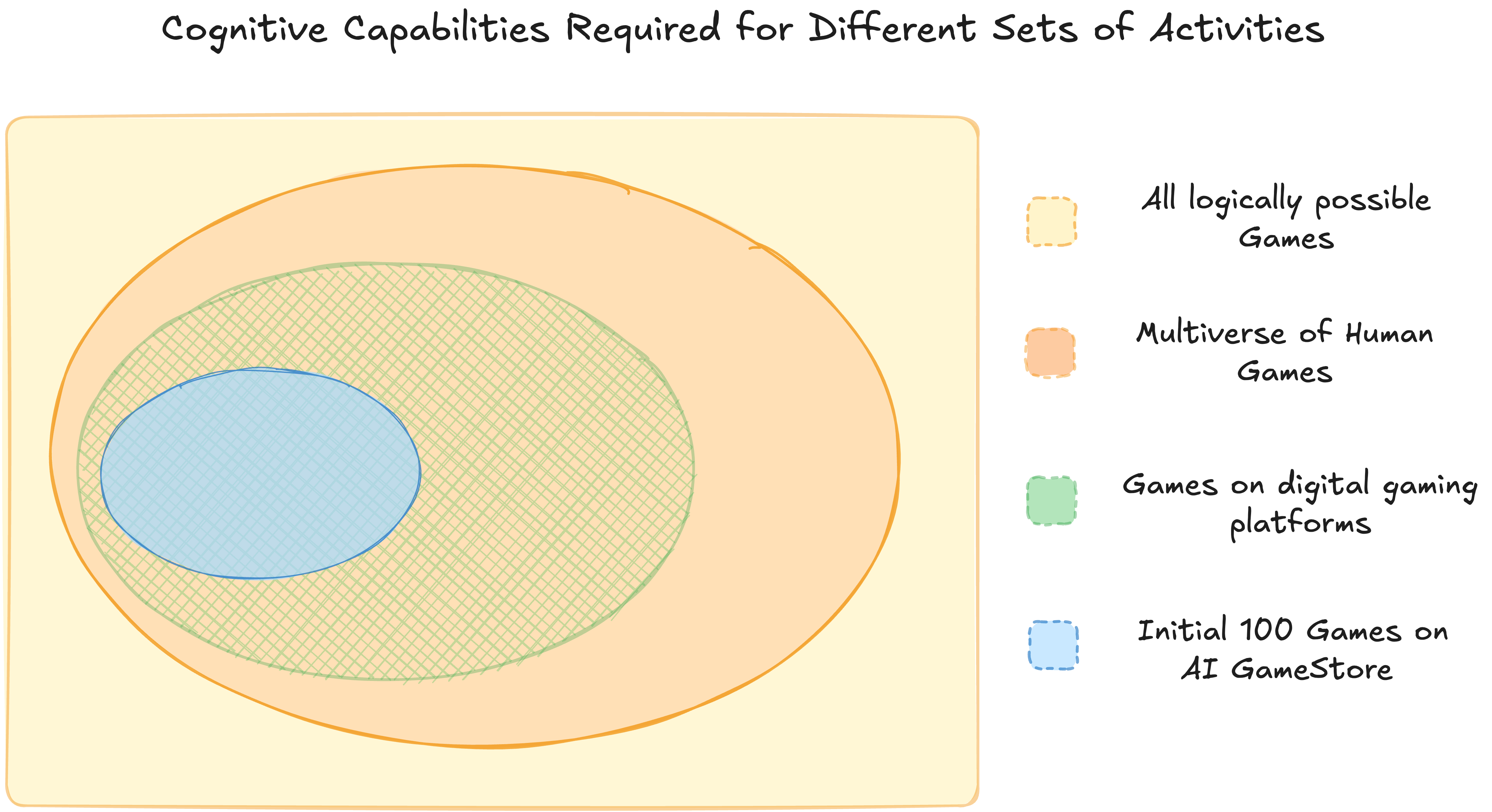}
    \caption{Comparison between different spaces of games. The yellow rectangle represents all computable games. The Multiverse of Human Games (orange) collectively demands cognitive capabilities found in average humans. The space of all digital games on gaming platforms (green) covers a subset. \textbf{\textsc{AI GameStore}} (blue) aims to sample from all digital games but the initial 100 games cover a restricted space.}
    \label{fig:venn_diagram}
\end{figure}

Figure \ref{fig:venn_diagram} illustrates the relationship between the different spaces of games discussed in this paper. The full space of all computable games is vast, but only a subset, what we refer to as the Multiverse of Human Games, consists of games that humans would create and enjoy. We argue that this space of games (orange) is a promising testbed for measuring progress towards human-like general intelligence. Digital games on gaming platforms cover a further subset, and the current \textbf{\textsc{AI GameStore}} samples from this space. Our initial 100 games are primarily casual digital games that can be learned quickly, and thus cover only a restricted region. Our goal is to continuously expand the complexity, quantity, and diversity of games to close the gap between the \textbf{\textsc{AI GameStore}} and the broader space of digital human games.

\section{Sourcing games}\label{sec:source_games}

For our first version of the \textbf{\textsc{AI GameStore}}, we first sourced from the Apple AppStore Top 100 games for each of 5 game categories (action, adventure, casual, puzzle, board) in 15 different countries (USA, China, UK, Japan, South Korea, South Africa, Mexico, India, France, Germany, Turkey, Saudi Arabia, Vietnam, Australia, Brazil). In total that resulted in 7500 games. In addition, we sourced top 500 Indie games from Steam (Steam does not maintain a per-country top chart).

\begin{figure}[h!]
    \centering
    \includegraphics[width=\linewidth]{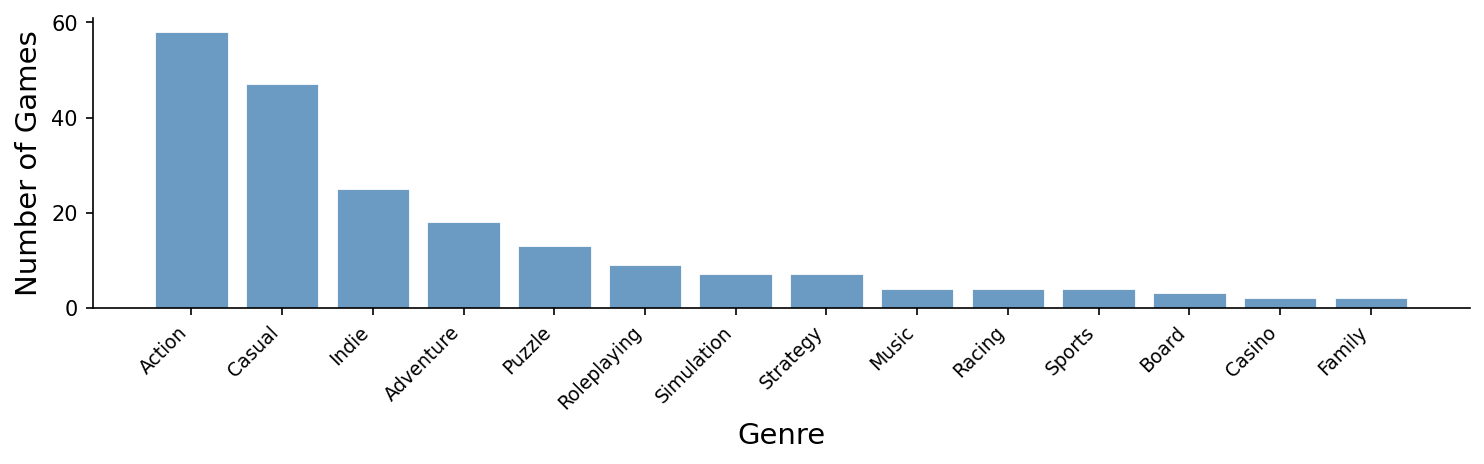}
    \caption{Genre distribution of the final curated dataset. The chart displays the frequency of games across 14 distinct game categories listed under the games. Action and Casual are the most common categories.}
    \label{fig:game_genre}
\end{figure}

The final set of 100 games show diverse genre, as shown in Figure \ref{fig:game_genre}. While action games represent the largest cohort, the evaluation suite also features many puzzle and board games.

\section{Game generation specification}\label{sec:game_specs}

\begin{itemize}
    \item We require all games written in JavaScript with \texttt{p5.js}. Games may use \texttt{three.js} and \texttt{matter.js} for 3D effects and physics simulations. This enable us to express a wide range of game mechanics with reasonably complex graphics. 
    
    \item We require all games to be playable exclusively through keyboard presses. While we envision future versions of \textbf{\textsc{AI GameStore}} to have games that require cursor movements, we restrict the output to key presses for the current \textbf{\textsc{AI GameStore}} games such that the action space of the AI model is more limited. This then becomes a multiple choice question for the model at each step. On the other hand, it is non-trivial to prompt models to use the cursor by outputting cursor trajectories. 
    
    \item All games can be paused and resumed. This enables us to query AI models for actions in real-time games. While we argue that a true generally intelligent model should be able to play the games as humans do, today's AI models have high latency for each API call, which limits the number of decisions they can make in each second.
    
    \item All games must include a scoring mechanism, and the scores should increase as the player makes progress over the game.
    \item All games should have multiple levels with a progressing degree of difficulty. This ensures that the game is neither too easy nor too hard, such that we can better quantify the performance of the players to inform comparison.
    \item For games where players can die or fail, there should be sufficient lives provided to allow players to learn and improve throughout the game session. Players can also reset the games at any moment in case the goal becomes no longer attainable.
\end{itemize}

\section{Game refinement}\label{sec:refine_interface}

The interface for generating and iteratively refining games is depicted in Figure 9. In the central screen, the game is displayed, allowing for real-time interaction. A human player can play the game in ``Human Mode'' or initiate automated testing. On the bottom right, the player can select a target LLM—such as Claude 4.5 Sonnet—to refine the game. The human player provides natural language feedback to describe specific issues or requested features in the ``Feedback'' text area, and click ``Apply Fix'' to trigger the automated code refinement, which will re-render the game for the player. The iteration and refinement process continues until the player is satisfied with the game. 

In our game generation, all game iteration and refinement was performed by the coauthors on this paper, but we believe this can be easily scaled up by recruiting online participants to play and refine the games as next steps for generating \textbf{\textsc{AI GameStore}} games.

\begin{figure}[h!]
    \centering
    \includegraphics[width=0.9\linewidth]{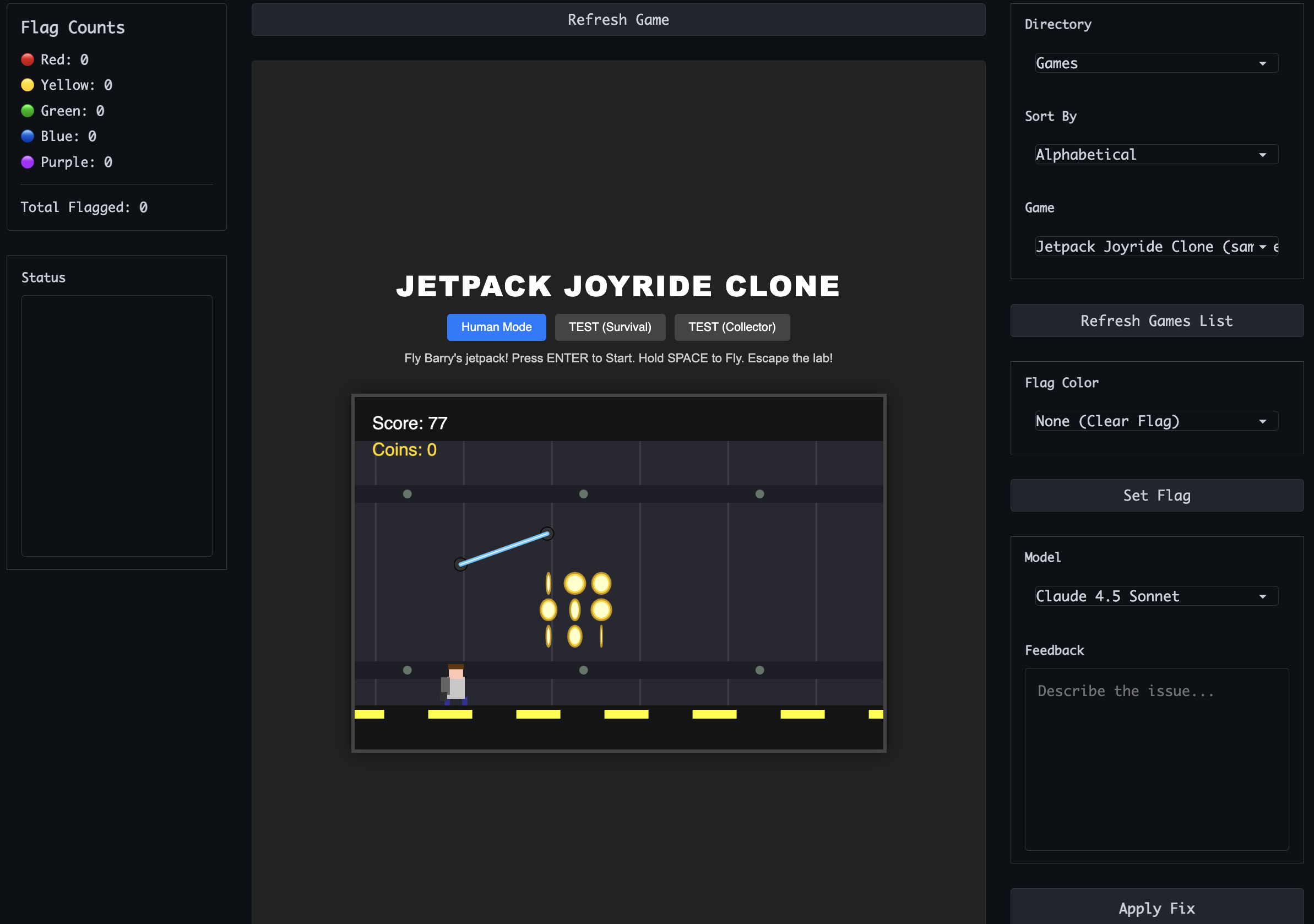}
    \caption{Interface for human-in-the-loop refinement}
    \label{fig:placeholder}
\end{figure}

\newpage

\section{Novel game-variant generation}\label{sec:novel_variants}

\begin{figure}[h!]
    \centering
    \includegraphics[width=0.99\linewidth]{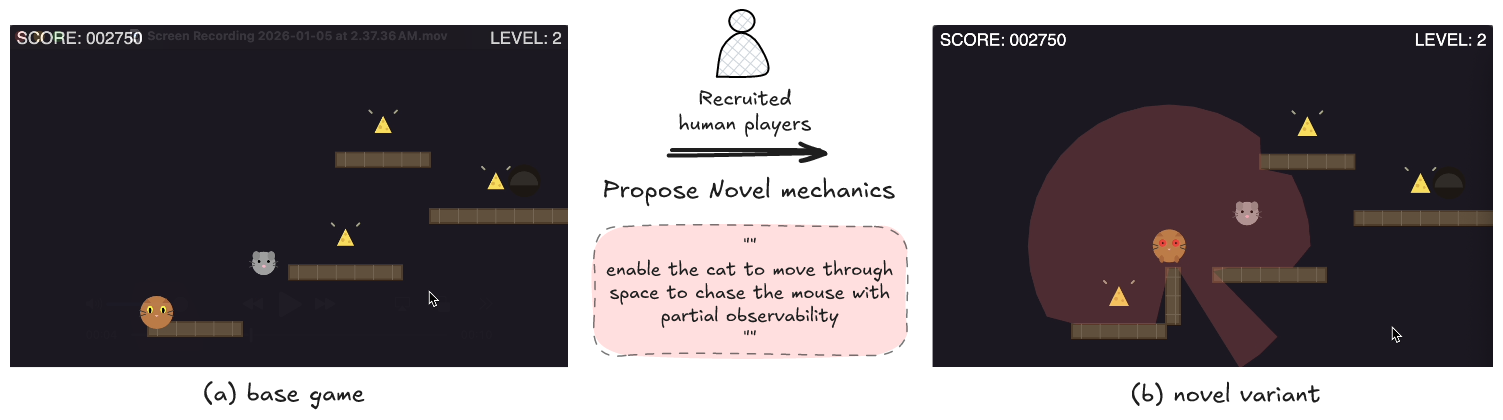}
    \caption{Generating novel variants of games. As a proof of concept, we recruited human players to play the base games and propose novel mechanics for variants in natural language, which is then implemented by an LLM to generate a novel game variant. We show an example above: the game objective is to control the mouse to get all the cheeses while avoiding the cat. In the base game, the cat moves on one of the platforms. In the novel variant, the cat can move through space to chase the mouse but has an isovist view. This enables us to quickly generate a large quantity of interesting games to evaluate AI models. At the same time, these novel variants enable us to control the demand profile of each game and stress test models. In the example above, the novel variant requires more sophisticated theory of mind reasoning and planning capabilities than the base game.}
    \label{fig:novel_variant}
\end{figure}

Multiple novel variants of the game can be generated by augmenting the game mechanics using the interface above. This allows the \textbf{\textsc{AI GameStore}} to generate a large quantity of test games from few source games. In addition, we can carefully control the cognitive capability demand for each variant by manipulating the mechanics, which allows for more targeted model evaluation. As a proof of concept, we recruited two human game players (both are undergraduate students who are not part of our author team) to play 30 of the base games and propose a novel variant for each. We show an example in Figure \ref{fig:novel_variant}. These games are not used in our model evaluation.

\section{Game annotation descriptions and rubrics} \label{sec:rubrics}

\begin{table}[h!]
\centering
\small
\begin{tabularx}{\textwidth}{@{} l X @{}}
\toprule
\textbf{Cognitive Capability} & \textbf{Description and Examples} \\ \midrule
\textbf{Visual Processing (VP)} & Requires counting or matching object properties like shape and size. \textit{Example:} \textit{Connect-3} games require identifying and matching tiles of the same color. \\ \addlinespace
\textbf{Spatial-temporal Coordination (ST)} & Requires well-timed and precise actions to navigate a visual scene. \textit{Example:} In \textit{Flappy Bird}, the player must time a sequence of flaps precisely to pass through barriers. \\ \addlinespace
\textbf{Memory (ME)} & Requires retrieving information from previous frames for current or future actions. \textit{Example:} Navigating a game with a partial map view requires integrating information across frames. \\ \addlinespace
\textbf{Planning (PL)} & Requires simulating many steps ahead and evaluating future outcomes. \textit{Example:} \textit{Water Sort} requires calculating a sequence of pours to reach the goal state. \\ \addlinespace
\textbf{World Model Learning (WM)} & Involves inferring hidden game mechanics not explicitly provided in the description through active gameplay. \textit{Example:} \textit{Baba Is You} requires players to discover how manipulating text blocks physically alters the game's logic and rules. \\ \addlinespace
\textbf{Physical Reasoning (PH)} & Requires mental physics simulation. \textit{Example:} \textit{Angry Birds} requires adjusting angles and power based on simulated trajectory and impact. \\ \addlinespace
\textbf{Social Reasoning (SO)} & Requires reasoning about the intentions, beliefs, and plans of other agents. \textit{Example:} In \textit{Counter Strike}, the player must reason about enemy's line of sight and pathing. \\
\bottomrule
\end{tabularx}
\vspace{0.2cm}
\caption{Cognitive capability categories for annotating \textbf{\textsc{AI GameStore}} games. Each capability is annotated on a 0-5 scale.}\label{tab:reasoning-categories}
\end{table}

In this section we show the rubrics for annotating the cognitive demands for each game. These rubrics are inspired by previous work by Zhou et al.~\citep{zhou2025general}.

The use of rubrics in natural language for annotating several capabilities of a task has several advantages: (1) It gives an interpretable, human-readable definition of the capability --the construct under consideration; (2) It extracts the demand levels for each game in a way that is independent from other games and from any player (human or machine); (3) It assumes no independence or any degree of correlation between the capabilities, allowing for flexibility in changes in the catalog of capabilities or observed correlations, without affecting the existing measures; and (4) It potentially allows for annotation automation \citep{zhou2025general} using multimodal LLMs.

\section*{Spatial-temporal coordination (ST)}
This criterion assesses the precision, timing, and sensorimotor integration necessary to navigate a visual scene and interact with dynamic elements. It measures the demand for real-time responsiveness and the synchronization of motor outputs with moving visual stimuli.

\noindent
\setcounter{savedtable}{\value{table}}\begin{xltabular}{\textwidth}{@{} L X @{}}
\toprule
\textbf{Level} & \textbf{Description and Example} \\
\midrule
None (0) & The task is static or turn-based. There is no timing demand, and performance remains unaffected by the speed of execution. \newline \textit{E.g., A digital version of Solitaire or a text-based RPG.} \\
\addlinespace
Very Low (1) & The task demands only basic, non-urgent interactions. Timing is extremely forgiving, and movement is slow, predictable, or linear. \newline \textit{E.g., Clicking on a large, slow-moving button that stays on screen for several seconds.} \\
\addlinespace
Low (2) & Reactive timing is necessary as the user must respond to a stimulus. Mistakes in timing are easily corrected and do not significantly impede overall progress. \newline \textit{E.g., Navigating a character through a wide corridor with no obstacles.} \\
\addlinespace
Intermediate (3) & Success depends on well-timed actions to navigate environments with moderate complexity. Performance relies on hitting specific ``windows'' of opportunity or maintaining a consistent rhythm. \newline \textit{E.g., Flappy Bird or basic platforming jumps.} \\
\addlinespace
High (4) & Multi-dimensional coordination is essential. The user must manage speed, trajectory, and timing simultaneously across multiple moving targets or obstacles. \newline \textit{E.g., A fast-paced ``bullet hell'' shooter or a 3D platformer needing mid-air adjustments.} \\
\addlinespace
Very High (5) & Frame-perfect precision and exceptional reflex integration are fundamental. Success depends on maintaining complex sequences of movement under high-velocity conditions where errors measured in milliseconds result in total failure. \newline \textit{E.g., Professional-level F1 racing simulations or high-level competitive fighting games.} \\
\bottomrule
\end{xltabular}\setcounter{table}{\value{savedtable}}

\vspace{2em}

\section*{Visual processing (VP)}
This criterion assesses the ability to identify, match, and categorize objects based on visual properties. It progresses from simple detection of presence to the complex parsing of cluttered, 3D, or partially occluded environments.

\noindent
\setcounter{savedtable}{\value{table}}\begin{xltabular}{\textwidth}{@{} L X @{}}
\toprule
\textbf{Level} & \textbf{Description and Example} \\
\midrule
None (0) & The task does not rely on visual differentiation. Information is conveyed through other modalities such as audio or text. \\
\addlinespace
Very Low (1) & The task involves detecting the presence or absence of one or few objects and their general location. \newline \textit{E.g., Detecting a white square on a black background.} \\
\addlinespace
Low (2) & Simple parsing of object identity and properties is sufficient. The user can distinguish between a few distinct shapes or colors, and high precision is not vital for success. \newline \textit{E.g., Sorting large, colored blocks into matching bins.} \\
\addlinespace
Intermediate (3) & The task involves identifying and matching multiple objects based on precise properties—such as shape, size, color, and texture—in a 2D environment. \newline \textit{E.g., Match-3 games like Bejeweled or Candy Crush.} \\
\addlinespace
High (4) & Successful navigation involves identifying objects despite occlusion, perspective shifts, or complex patterns in a crowded environment. \newline \textit{E.g., A ``Hidden Object'' game with significant clutter and overlapping items.} \\
\addlinespace
Very High (5) & High-level visual inference is necessary. The user must parse complex, 3D, or partially observable scenes to make educated guesses about object identity and position using minimal visual cues. \newline \textit{E.g., Identifying an enemy's position in a tactical shooter by seeing only a sliver of a shadow.} \\
\bottomrule
\end{xltabular}\setcounter{table}{\value{savedtable}}

\vspace{2em}

\section*{Memory (ME)}
This criterion assesses the demand for retrieving and integrating information from previous states to inform current or future actions. It spans from immediate sensory persistence to the long-term synthesis of vast amounts of data.

\noindent
\setcounter{savedtable}{\value{table}}\begin{xltabular}{\textwidth}{@{} L X @{}}
\toprule
\textbf{Level} & \textbf{Description and Example} \\
\midrule
None (0) & The task is entirely reactive. All information needed to succeed is present in the current view or frame. \newline \textit{E.g., A simple reaction test where you click when a light turns green.} \\
\addlinespace
Very Low (1) & Successful completion involves holding basic information from the immediate past to inform future actions. However, this information is not critical for the success of the player. \newline \textit{E.g., Remembering which direction a character was just walking to continue the path.} \\
\addlinespace
Low (2) & Remembering multiple pieces of information for a short duration, often a few seconds, is necessary. \newline \textit{E.g., A standard ``Match Two'' card memory game.} \\
\addlinespace
Intermediate (3) & The task involves integrating information across a longer horizon, such as building a mental map of a small local area from exploration. \newline \textit{E.g., Navigating a maze where only the immediate surroundings are visible (Fog of War).} \\
\addlinespace
High (4) & Tracking multiple types of information across long durations—such as inventory states, quest locations, and character status—is essential. \newline \textit{E.g., Keeping track of multiple resource counts and enemy positions in a Real-Time Strategy (RTS) game.} \\
\addlinespace
Very High (5) & The retention and synthesis of vast, heterogeneous datasets across long timeframes are vital. This often involves recalling specific narrative details or mechanics encountered prior to solving current problems. \newline \textit{E.g., An RPG where information from hours before affects the current decision-making.} \\
\bottomrule
\end{xltabular}\setcounter{table}{\value{savedtable}}

\vspace{2em}

\section*{World-model learning (WM)}
This criterion assesses the ability to infer hidden mechanics, rules, or causal relationships through active experimentation and hypothesis testing.

\noindent
\setcounter{savedtable}{\value{table}}\begin{xltabular}{\textwidth}{@{} L X @{}}
\toprule
\textbf{Level} & \textbf{Description and Example} \\
\midrule
None (0) & All rules and mechanics are explicitly stated, standardized, or immediately obvious. No discovery is necessary. \newline \textit{E.g., A standard Tic-Tac-Toe game.} \\
\addlinespace
Very Low (1) & The environment is highly familiar. Minimal trial-and-error is needed to understand basic mechanics, leaving little uncertainty about how the system works. \newline \textit{E.g., A basic maze game with multiple colored exits.} \\
\addlinespace
Low (2) & The environment contains a few novel mechanics that can be easily discovered through casual interaction. \newline \textit{E.g., Learning that a specific type of button can unlock certain doors.} \\
\addlinespace
Intermediate (3) & Success involves inferring ``hidden'' mechanics within a limited search space. The user must use simple trial-and-error to understand how different elements interact. \newline \textit{E.g., Learning to unlock a door in a particular, unusual way.} \\
\addlinespace
High (4) & The task presents a novel system, prompting the user to discover unfamiliar mechanics through deliberate exploration and observation of patterns. \newline \textit{E.g., Understanding complex crafting recipes in a survival game without a manual.} \\
\addlinespace
Very High (5) & The environment is highly novel and potentially counter-intuitive. Success depends on setting up ``epistemic goals''—deliberate experiments to validate or invalidate hypotheses about how the world works. \newline \textit{E.g., Games like ``Outer Wilds,'' where the player must scientifically deduce physical laws to progress.} \\
\bottomrule
\end{xltabular}\setcounter{table}{\value{savedtable}}

\vspace{2em}

\section*{Planning (PL)}
This criterion assesses the requirement for simulating future states and evaluating the outcomes of a sequence of actions. It measures the depth of the ``search tree'' and the complexity of managing branching possibilities.

\noindent
\setcounter{savedtable}{\value{table}}\begin{xltabular}{\textwidth}{@{} L X @{}}
\toprule
\textbf{Level} & \textbf{Description and Example} \\
\midrule
None (0) & Actions are purely reactive or instinctive, as there is no advantage to thinking ahead. \newline \textit{E.g., Whack-a-Mole.} \\
\addlinespace
Very Low (1) & Thinking exactly one step ahead is necessary to avoid immediate negative outcomes. \newline \textit{E.g., Moving out of the way of a slow-moving projectile.} \\
\addlinespace
Low (2) & Planning short, linear sequences (2–3 steps) with a clear, singular goal is necessary for progress. \newline \textit{E.g., Moving a piece in Checkers to set up a single jump.} \\
\addlinespace
Intermediate (3) & Simulating several steps ahead and evaluating a few possible future outcomes are necessary to reach a specific goal state. \newline \textit{E.g., Solving a ``Water Sort'' puzzle or a medium-difficulty Sudoku.} \\
\addlinespace
High (4) & Deep search is a fundamental requirement. The player must account for multiple branching possibilities and plan many moves into the future, often anticipating an opponent's counter-moves. \newline \textit{E.g., High-level Chess or Go.} \\
\addlinespace
Very High (5) & Strategic, multi-objective planning in dynamic or stochastic environments is essential. The user must balance long-term goals against immediate threats while accounting for uncertainty. \newline \textit{E.g., Grand Strategy games where economic, military, and diplomatic plans are managed simultaneously.} \\
\bottomrule
\end{xltabular}\setcounter{table}{\value{savedtable}}

\vspace{2em}

\section*{Physical reasoning (PH)}
This criterion assesses the mental simulation of physical properties, such as gravity, trajectory, momentum, and material interactions. It measures the ability to predict how objects will behave according to physical laws.

\noindent
\setcounter{savedtable}{\value{table}}\begin{xltabular}{\textwidth}{@{} L X @{}}
\toprule
\textbf{Level} & \textbf{Description and Example} \\
\midrule
None (0) & The task follows abstract or symbolic rules with no physical components. \newline \textit{E.g., A crossword puzzle or a math quiz.} \\
\addlinespace
Very Low (1) & Basic awareness of physical laws such as ``solidity'' is sufficient for the task (e.g., characters cannot walk through walls). \newline \textit{E.g., Navigating a top-down RPG where walls block movement.} \\
\addlinespace
Low (2) & Understanding simple linear movement or basic gravity is necessary. \newline \textit{E.g., Dropping an object and knowing it will fall straight down.} \\
\addlinespace
Intermediate (3) & The task requires simulating object trajectories. The user must make precise predictions of how an object will move through space to interact with other objects. \newline \textit{E.g., Simple Angry Birds levels with few structures} \\
\addlinespace
High (4) & Reasoning about complex physical interactions with few basic variables—such as leverage, friction, momentum transfer, or basic fluid dynamics—is necessary. \newline \textit{E.g., Angry Birds with complex structures that may interact with each other} \\
\addlinespace
Very High (5) & The simultaneous integration of multiple physical variables (wind, mass, elasticity, torque) is necessary to predict outcomes in highly dynamic environments. \newline \textit{E.g., Building and flying a rocket in ``Kerbal Space Program.''} \\
\bottomrule
\end{xltabular}\setcounter{table}{\value{savedtable}}

\vspace{2em}

\section*{Social reasoning (SO)}
This criterion assesses the cognitive demands associated with mind modeling and social cognition. The levels progress from tasks needing no mind modeling to those requiring reasoning about how the beliefs, desires, and emotions of multiple agents interact.

\noindent
\setcounter{savedtable}{\value{table}}\begin{xltabular}{\textwidth}{@{} L X @{}}
\toprule
\textbf{Level} & \textbf{Description and Example} \\
\midrule
None (0) & The task does not involve mind modeling or social cognition. It may not involve other agents, or if it does, interacting with them is not necessary for success. \newline \textit{E.g., Solving a Sudoku puzzle.} \\
\addlinespace
Very Low (1) & Performance is improved through the detection of other agents. These agents often don't move or act in a goal-directed way (e.g. periodic movements). Reasoning about observed behavior or attributing mental states to others is not necessary for good performance. \newline \textit{E.g. a platformer game where an enemy moves horizontally on a particular platform.} \\
\addlinespace
Low (2) & This task requires some basic intuition about the behaviour of others, but only minimal levels of mental state attribution. Good performance might be based on developing accurate associations between other’s responses and the stimuli that caused them. Note, this reasoning need not be explicit. \\
\addlinespace
Intermediate (3) & This task goes beyond simple state-behaviour associations and involves attributing cognitive or affective states (i.e., mentalising). That is, it involves inferring and representing specific mental properties about others (‘they believe the moon landing was a hoax’, ‘they want a glass of water’). The task may not, however, require explicit reasoning about these mental states (i.e., full-blown theory of mind). \newline \textit{E.g. Recognizing that someone using a rock to crack open a coconut is trying to get to the food inside.} \\
\addlinespace
High (4) & This task requires a full theory of mind to be solved effectively. It requires not only the attribution of mental states to others, but explicit reasoning about those states. It may also require the integration of social knowledge and heuristics about normal agentic behaviour to accurately predict future behaviour. Importantly, this task also requires a clear distinction between self- and other-related representations. \newline \textit{E.g. A typical false belief task (e.g. Sally Anne test)} \\
\addlinespace
Very High (5) & This task requires exceptional mind modelling and social cognition abilities. It goes beyond generating intuitive theories about another agent within a dyadic interaction, and instead requires the combination of multiple theories of mind corresponding to the intentions, emotions, and beliefs of a range of different agents. Expanding the scope of mind-modelling and social cognition to include multiple agents would enable more sophisticated forms of collaborative action. Tasks at this level may require an understanding of the complex networks and hierarchies that form within social groups \newline \textit{E.g. Leading a negotiation between multiple stakeholders where each party has different beliefs about others' intentions and bottom lines, while managing the complex emotional dynamics between opposing personalities.} \\
\bottomrule
\end{xltabular}\setcounter{table}{\value{savedtable}}

\section{Comparing public and all games} \label{sec:public-games}

We evaluate the representativeness of the 10 publicly released games against the full 100-game dataset. Overall, the comparison shows that the public subset serves as a high-fidelity proxy for the broader benchmark, as evidenced by the comparable mean ratings for human-centric metrics such as "Funness" and "Challengingness" as well as model performance.

\begin{xltabular}{\textwidth}{@{} p{2.5in} X X @{} }

    \toprule
    \textbf{Metric / Model} & \textbf{10 Public Games} & \textbf{All 100 Games} \\ 
    \midrule
    \endfirsthead
    
    \multicolumn{3}{c}{} \\
    \toprule
    \textbf{Metric / Model} & \textbf{10 Public Games} & \textbf{All 100 Games} \\ 
    \midrule
    \endhead

    \textbf{Participant Ratings (Mean $\pm$ SD)} & & \\
    \quad Average Funness Rating & 51.41 $\pm$ 9.54 & 52.37 $\pm$ 13.17 \\
    \quad Average Challengingness Rating & 56.32 $\pm$ 7.69 & 59.33 $\pm$ 13.89 \\
\midrule
   \multicolumn{3}{@{} l}{\textbf{Model Performance (Geom Mean [95\% CI])}} \\
    \quad GPT-5.2 & 6.71 [2.67, 16.22] & 8.26 [5.93, 11.28] \\
    \quad Claude-Opus-4.5 & 5.91 [2.06, 16.90] & 7.74 [5.50, 10.68] \\
    \quad Gemini-2.5-Pro & 8.99 [4.08, 18.87] & 7.49 [5.36, 10.28] \\
    \quad Gemini-2.5-Flash & 6.27 [2.27, 14.05] & 7.07 [5.07, 9.69] \\
    \quad GPT-5-mini & 5.01 [1.67, 13.92] & 6.13 [4.44, 8.39] \\
    \quad Llama-4-Maverick & 4.74 [2.22, 9.12] & 5.91 [4.26, 7.80] \\
    \quad Qwen-3-VL-32B & 2.53 [1.16, 5.97] & 4.68 [3.39, 6.41] \\
    \midrule
    \textbf{Cognitive Demands (Mean $\pm$ SD)} & & \\
    \quad Visual Processing (VP) & 2.40 $\pm$ 0.70 & 2.23 $\pm$ 0.63 \\
    \quad Spatial Temporal Coordination (ST) & 2.10 $\pm$ 1.52 & 2.04 $\pm$ 1.54 \\
    \quad Memory (ME) & 0.60 $\pm$ 0.97 & 0.20 $\pm$ 0.72 \\
    \quad Planning (PL) & 2.20 $\pm$ 1.23 & 0.95 $\pm$ 1.17 \\
    \quad Word Model Learning (WM) & 0.80 $\pm$ 1.32 & 0.52 $\pm$ 1.06 \\
    \quad Physical Reasoning (PH) & 1.30 $\pm$ 0.95 & 0.96 $\pm$ 1.20 \\
    \quad Social Reasoning (SO) & 0.90 $\pm$ 0.99 & 0.50 $\pm$ 0.81 \\
    \bottomrule

    \caption{Comparing the 10 public games vs all 100 games on the \textbf{\textsc{AI GameStore}}. Values for Model Performance are scaled by 100 to represent the percentage relative to the human median (100).} \label{tab:game_comparison} \\
\end{xltabular}

\section{Model experiment details} \label{sec:harness}

In each API call, the model is prompted to give a list of actions to perform. The model is given a list of possible keys to choose from. For each api call, the model is instructed to return a list of 5 objects, where each object is a list of action strings. Each object corresponds to the agent actions for a 0.2 second period. The list of action strings indicate which keys are pressed for this 0.2 second period. If multiple action strings are included, then all mentioned keys are pressed. For each key press, the model can also indicate a regular key press ("DOWN") or a hold key ("HOLD\_DOWN"). A regular key press indicates that the action is applied once, whereas a HOLD option would indicate the action is continuously applied for the whole 0.2 second period. The model can always return "RETRY" to restart the level.

\begin{figure*}[h!]
    \includegraphics[width=\textwidth]{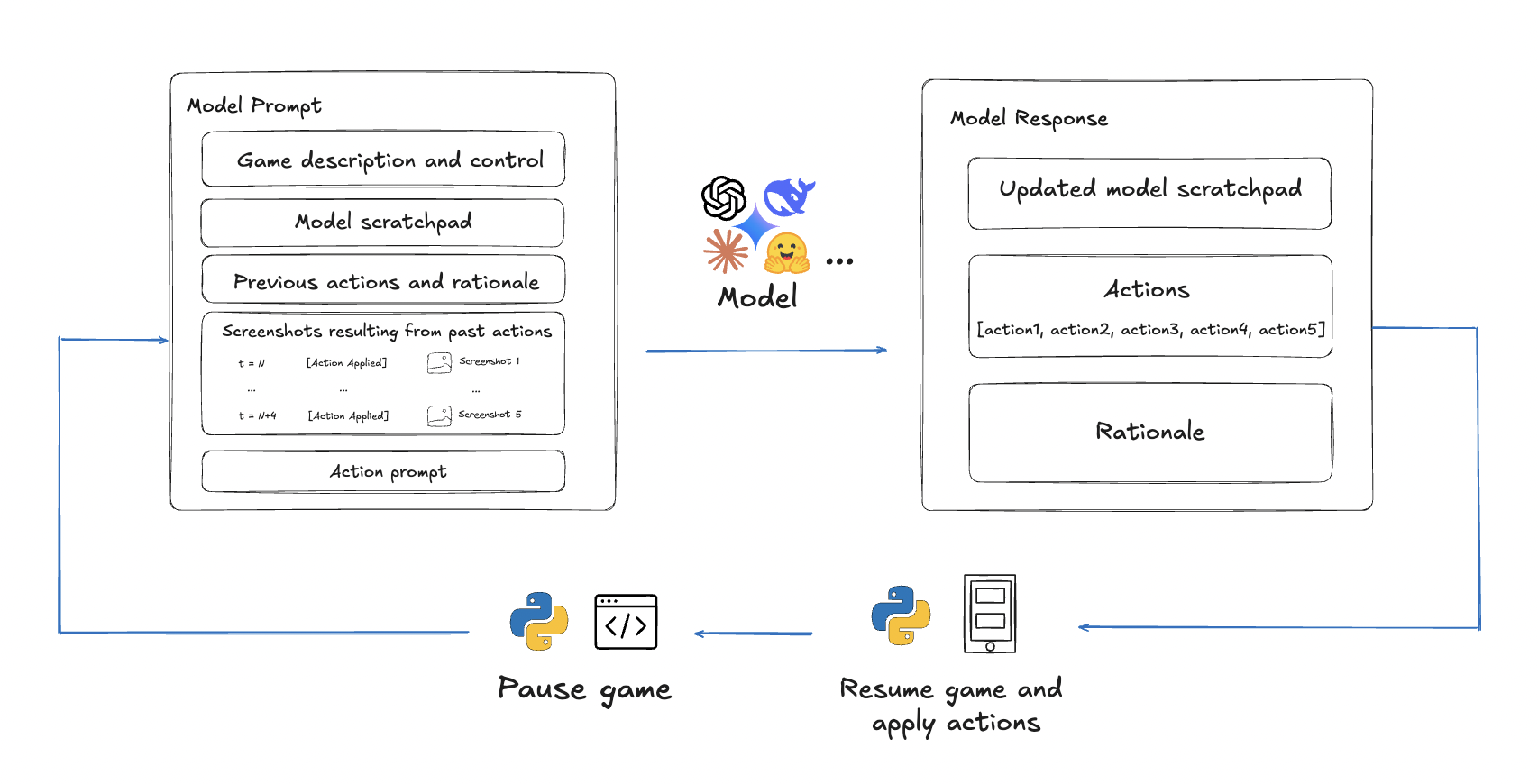}
    \caption{The model evaluation harness for evaluating AI models on \textbf{\textsc{AI GameStore}} games. The Model Prompt provides the AI with the current game state via screenshots, along with game descriptions, a "scratchpad" for maintaining internal state, and a history of previous actions and rationales. The model then processes this information to generate a Model Response, which includes an updated scratchpad, a determined sequence of actions, and a rationale for those choices. These actions are then applied to the game environment, and the game is paused to capture the new state, completing the cycle for the next interaction step.}
    \label{harness}
\end{figure*}

\begin{tcolorbox}[
    breakable,              %
    width=\textwidth,       %
    boxsep=0pt,
    left=5pt,               %
    right=5pt,
    top=5pt,
    bottom=5pt,
    arc=0pt,
    boxrule=0.5pt,
    toprule=2pt,
    colback=white,
    title={\textbf{Prompt for VLM actions at each gameplay step}},
    colbacktitle=gray!20,   %
    coltitle=black          %
]

You are a professional video game player tasked to win a video game. You will read the description of the game and your previous actions and game state. You will then provide actions for the next 5 steps (Each step lasts for 0.2 seconds).

[INSERT GAME DESCRIPTION, SCRATCHPAD, PREVIOUS ACTIONS and SCREENSHOTS]

\begin{lstlisting}[breaklines]

**Output:**

1. Provide a brief reasoning behind your actions (< 10 sentences).

2. Output exactly 5 lists of actions. Each list represents a 0.2 second time segment.
   - Each segment can contain: [NOOP] (do nothing), a single action like [UP], or multiple simultaneous actions like [UP, LEFT]
   - Instant actions (applied once at the start of the segment): "UP", "DOWN", "LEFT", "RIGHT", "SPACE"
   - Continuous actions (held for the entire 0.2 seconds): "HOLD_UP", "HOLD_DOWN", "HOLD_LEFT", "HOLD_RIGHT", "HOLD_SPACE"
   - You can mix instant and continuous actions in the same segment, e.g., [UP, HOLD_LEFT] applies UP once and holds LEFT for 0.2s
   - You can use "R" to restart the game if it ends. Feel free to restart as many time as you want.

**Format your response as follows:**
<reasoning>
[INSERT YOUR THINKING]
</reasoning>
<keys>
[["NOOP"], ["HOLD_UP", "HOLD_LEFT"], ["NOOP"], ["HOLD_UP"], ["DOWN"]]
</keys>
<scratchpad>
Provide a scratchpad of your current understanding of the game state, your plan, and any important observations. This will be included in future API calls to help maintain context.
</scratchpad>

\end{lstlisting}
\end{tcolorbox}

\section{Additional experimental results} \label{sec:additional_results}

The model median normalized score and geometric mean score across all 100 games are shown in Table \ref{tab:median_vs_mean}.

\begin{table}[h]
    \centering
    \begin{tabular}{@{}lcc@{}}
        \toprule
        \textbf{Model} & \textbf{Median [95\% CI]} & \textbf{Geom Mean [95\% CI]} \\ \midrule
        Gemini-2.5-Pro    & 12.40 [4.78, 18.10] & 8.99 [4.08, 18.87] \\
        GPT-5.2           & 10.70 [5.57, 17.09] & 6.71 [2.67, 16.22] \\
        Claude-Opus-4.5   & 8.63 [4.02, 17.86]  & 5.91 [2.06, 16.90] \\
        Gemini-2.5-Flash  & 7.32 [4.03, 13.72]  & 6.27 [2.27, 14.05] \\
        Llama-4-Maverick  & 6.50 [3.94, 11.31]  & 4.74 [2.22, 9.12]  \\
        GPT-5-mini        & 6.36 [2.01, 14.97]  & 5.01 [1.67, 13.92] \\
        Qwen-3-VL-32B     & 3.50 [0.00, 7.41]   & 2.53 [1.16, 5.97]  \\ \bottomrule
    \end{tabular}
    \vspace{1em}
    \caption{Model median vs geometric mean performance on all 100 scores.}
    \label{tab:median_vs_mean}
\end{table}

\begin{figure}[h!]
    \centering
    \includegraphics[width=\linewidth]{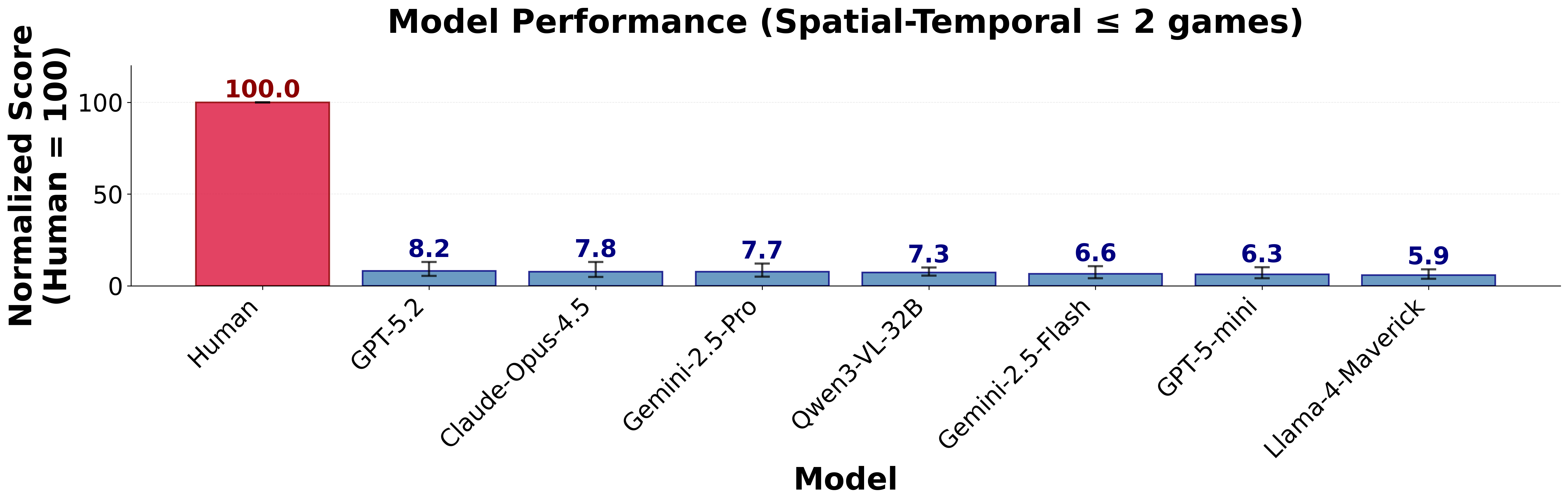}
    \caption{Model performance on games that require low spatial temporal coordination. We find that the results are not significantly different from the aggregate model performance.}
    \label{fig:low_spatial_temporal}
\end{figure}

Figure \ref{fig:low_spatial_temporal} show models' performance on games that require low spatio-temporal coordination (demand score $\leq$ 2) in Figure \ref{fig:low_spatial_temporal}. We did not observe any significant changes to top models' performance (e.g. \textsc{GPT-5.2} or \textsc{Gemini-2.5-pro}). This indicates that model failure is not solely due to slow reaction time for games.

\begin{figure*}[h!]
    \centering
    \includegraphics[width=\linewidth]{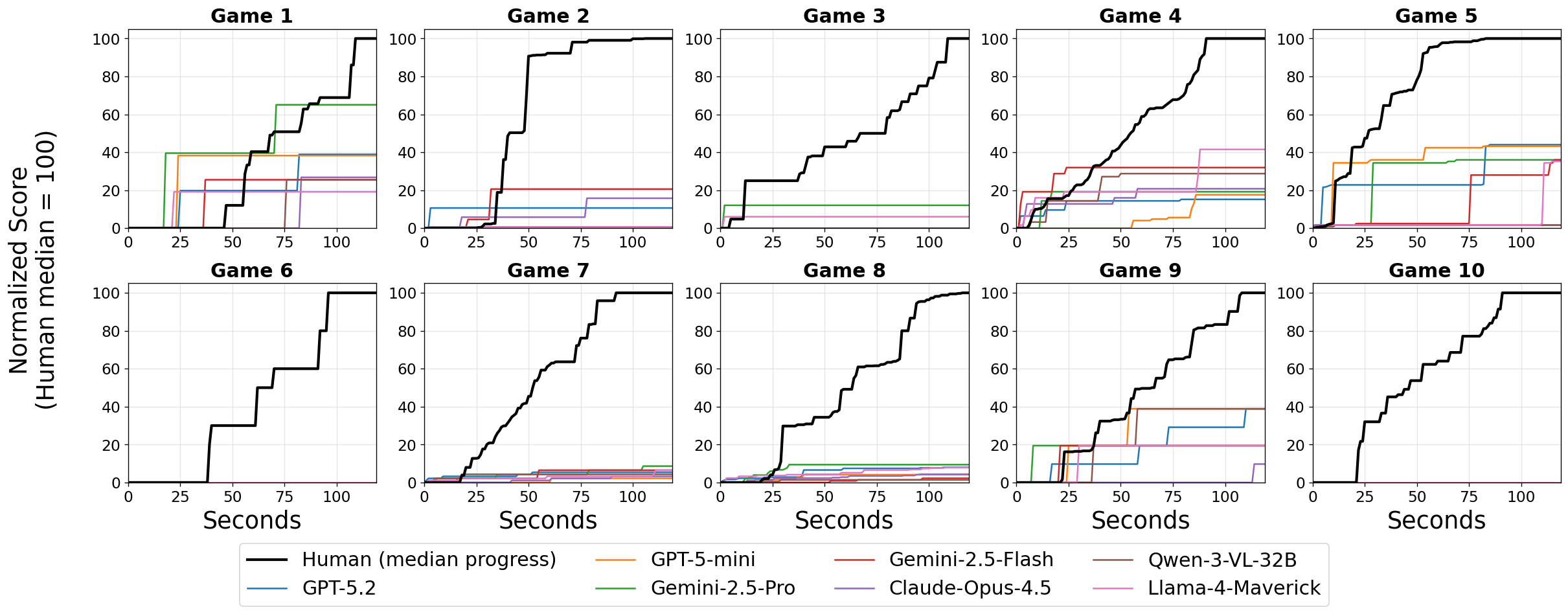}
    \caption{Model and human median cumulative score trajectories on 10 public games. Scores are based on the maximum achieved up to each time point, measured each second for 120s, normalized to human median = 100. Models typically make some progress but much more slowly than humans, and completely fail on a number of games.}
    \label{fig:model_vs_human}
\end{figure*}

\clearpage
\section{Benchmark validation, coverage, and protocol sensitivity}\label{sec:benchmark_validation}

This appendix reports the additional analyses introduced in response to reviewer questions about construct validity, representativeness, annotation reliability, and evaluation protocol. These analyses support the benchmark's intended use while making explicit the limits of the present 100-game release.

\subsection{Generated-game fidelity}\label{sec:fidelity_study}

We recruited 25 Prolific participants for a perceptual and mechanical fidelity study on the ten public games. After playing an adapted game for two minutes, each participant watched a 30-second clip from the source game and a randomly selected non-source game, then rated visual similarity, mechanics similarity, and cognitive-capability match on 0--100 sliders. As Table \ref{tab:fidelity_study} shows, participants rated the source clips substantially higher on all three dimensions. This study tests whether the adaptations preserve recognizable task structure; it does not imply pixel-level reproduction or equivalence in every design detail.

\begin{table}[H]
\centering
\small
\begin{tabular}{@{}lccc@{}}
\toprule
\textbf{Rating} & \textbf{Source} & \textbf{Non-source} & \textbf{Difference} \\
\midrule
Visual similarity & $79.7 \pm 2.0$ & $11.5 \pm 1.7$ & $68.2$ \\
Mechanics similarity & $85.6 \pm 1.9$ & $16.8 \pm 2.2$ & $68.8$ \\
Cognitive-capability match & $80.6 \pm 2.5$ & $29.9 \pm 2.9$ & $50.8$ \\
\bottomrule
\end{tabular}
\caption{Generated-game fidelity ratings (mean $\pm$ standard error) from 25 participants. Higher ratings indicate greater similarity between an adapted game and the comparison clip.}
\label{tab:fidelity_study}
\end{table}

\subsection{Convergent validity}\label{sec:convergent_validity}

To test whether capability-specific rankings align with established measures, we computed Spearman rank correlations across the seven evaluated models between \textbf{\textsc{AI GameStore}} capability scores and conceptually related benchmarks. Table \ref{tab:convergent_validity} reports strong positive associations for all six capabilities for which a defensible static comparison was available. We omit spatial--temporal coordination because the candidate static benchmarks do not measure real-time control. The overall \textbf{\textsc{AI GameStore}} ranking also correlates with the Artificial Analysis Intelligence Index ($\rho=.94$, $p=.002$; Pearson $r=.81$). With only seven models, these values should be interpreted as preliminary convergent evidence rather than precise population estimates.

\begin{table}[H]
\centering
\small
\begin{tabular}{@{}llc@{}}
\toprule
\textbf{AI GameStore capability} & \textbf{External benchmark} & \textbf{Spearman $\rho$} \\
\midrule
Physical reasoning & PIQA & $0.90$ \\
Visual processing & MMMU & $0.89$ \\
Social reasoning & MMToM-QA & $0.82$ \\
Planning & Blocksworld & $0.82$ \\
Memory & BABILong & $0.70$ \\
World-model learning & ARC-AGI-1 & $0.66$ \\
\bottomrule
\end{tabular}
\caption{Convergent validity of capability-specific model rankings across seven models.}
\label{tab:convergent_validity}
\end{table}

\subsection{Coverage in the broader game space}\label{sec:coverage_analysis}

\begin{figure}[H]
\centering
\includegraphics[width=0.96\linewidth]{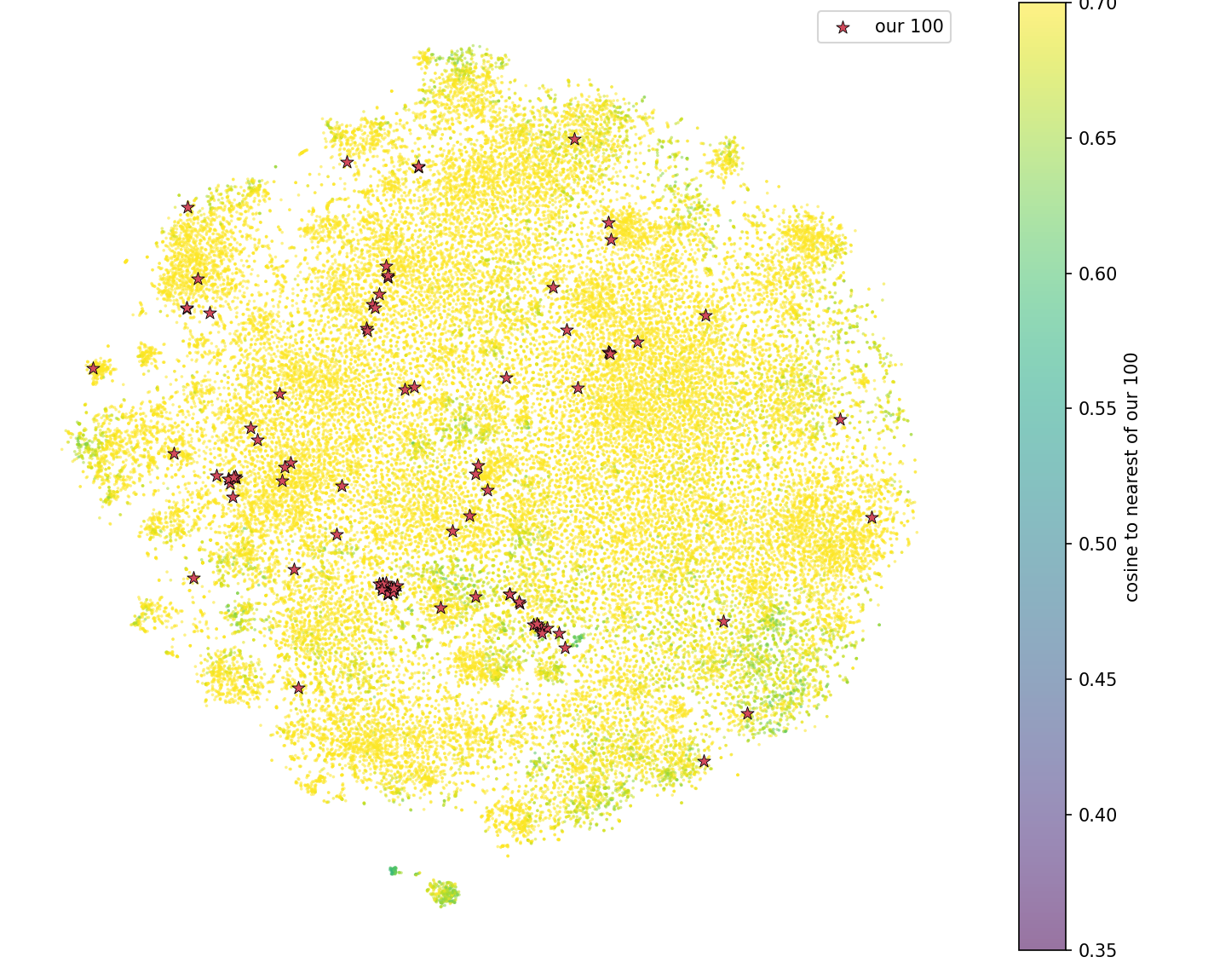}
\caption{Embedding coverage of 100 \textbf{\textsc{AI GameStore}} descriptions relative to approximately 345,000 Steam and IGDB titles. Red stars denote benchmark games; catalog points are colored by cosine similarity to their nearest benchmark game. The two-dimensional t-SNE projection is exploratory and should not be interpreted as preserving all global distances.}
\label{fig:embedding_coverage}
\end{figure}

We jointly embedded descriptions of the 100 \textbf{\textsc{AI GameStore}} games with approximately 345,000 Steam and IGDB titles using Nomic-embed-text-v1.5, then visualized the joint space with t-SNE (Figure \ref{fig:embedding_coverage}). The benchmark games appear across many regions of the catalog rather than in a single isolated cluster. Their nearest catalog neighbors span 29 of 31 genre families containing at least 1,000 titles. Narrative, simulation, real-time strategy, and quiz/trivia are comparatively under-represented and are priorities for expansion. Because t-SNE distorts global distances, the plot is descriptive evidence of local semantic coverage, not proof that the current suite represents the full game distribution.

\subsection{Suite composition and remaining gaps}\label{sec:suite_composition}

All 100 current games use keyboard controls and a single two-minute episode, so long-horizon interaction is not yet represented. Forty-five games are real-time or reaction-heavy, while 55 are turn-based or puzzle-like with little time pressure. Using the highest annotated capability score as the dominant-demand label, 70 games have a single dominant capability and 30 are balanced across multiple capabilities (Table \ref{tab:suite_composition}). The release therefore emphasizes spatial--temporal and visual demands and contains no game whose sole dominant label is memory, even though memory is often required jointly with other capabilities.

\begin{table}[H]
\centering
\small
\begin{tabular}{@{}lr@{}}
\toprule
\textbf{Suite category} & \textbf{Games} \\
\midrule
Real-time or reaction-heavy & 45 \\
Turn-based or puzzle-like & 55 \\
\midrule
Single dominant capability & 70 \\
\quad Spatial--temporal coordination & 39 \\
\quad Visual processing & 17 \\
\quad Physical reasoning & 6 \\
\quad Planning & 4 \\
\quad World-model learning & 3 \\
\quad Social reasoning & 1 \\
\quad Memory & 0 \\
Balanced across capabilities & 30 \\
\bottomrule
\end{tabular}
\caption{Composition of the initial 100-game release. Capability counts describe the 70 games with a single dominant demand; the remaining 30 are balanced.}
\label{tab:suite_composition}
\end{table}

Grouping model performance by dominant demand reproduces the difficulty ordering in the main results (Table \ref{tab:dominant_performance}). Games dominated by social reasoning, planning, and world-model learning are hardest for models, whereas visual- and spatial--temporal-dominant games yield the highest scores. Because several capabilities dominate only a handful of games and most games draw on multiple capabilities, a model may still fail a game because of a non-dominant demand; these group means are therefore descriptive. With only 100 games and sparse coverage of some capabilities, a regression of performance on the full demand profile was not robust, which motivates targeted sourcing of games in sparse regions of the capability space.

\begin{table}[H]
\centering
\small
\begin{tabular}{@{}lrr@{}}
\toprule
\textbf{Dominant capability} & \textbf{$n$} & \textbf{Model geomean (\% of human)} \\
\midrule
Social reasoning & 1 & 3.39 \\
Planning & 4 & 3.59 \\
World-model learning & 3 & 4.24 \\
Physical reasoning & 6 & 6.94 \\
Spatial--temporal coordination & 38 & 8.33 \\
Visual processing & 17 & 15.51 \\
\midrule
Balanced (no dominant capability) & 29 & 4.72 \\
\bottomrule
\end{tabular}
\caption{Model geometric-mean score (percentage of human performance) grouped by each game's dominant cognitive demand. Lower values indicate games that are harder for models.}
\label{tab:dominant_performance}
\end{table}

\subsection{Annotation reliability}\label{sec:annotation_reliability}

Three author annotators independently scored every game before deliberation using the behaviorally anchored 0--5 rubrics in Appendix \ref{sec:rubrics}. Squared-distance Krippendorff's $\alpha$ was $.81$ overall and ranged from $.56$ to $.84$ across individual capabilities; mean pairwise quadratic-weighted Cohen's $\kappa$ was $.81$. Annotators then reviewed disagreements and produced the adjudicated profiles used in the main analyses. The range across capabilities suggests that future releases should continue refining anchors for the least reliable dimensions.

\subsection{Protocol sensitivity, compute, and scaffold effects}\label{sec:protocol_sensitivity}

Table \ref{tab:token_usage} reports median token use per game for the canonical model runs. Token count complements wall-clock runtime because provider latency varies, although tokens are not perfectly comparable across proprietary tokenizers or reasoning implementations. The 0.2-second action segments approximate the scale of human visual response, and restricting the analysis to games with spatial--temporal demand $\leq 2$ leaves model geometric means between 5.9 and 8.2 (Figure \ref{fig:low_spatial_temporal}), so slow reaction control alone does not explain the performance gap.

\begin{table}[H]
\centering
\small
\begin{tabular}{@{}lr@{}}
\toprule
\textbf{Model} & \textbf{Median tokens per game} \\
\midrule
GPT-5.2 & 108,285 \\
Claude-Opus-4.5 & 159,845 \\
Gemini-2.5-Pro & 285,091 \\
Gemini-2.5-Flash & 382,148 \\
GPT-5-mini & 201,908 \\
Llama-4-Maverick & 145,652 \\
Qwen-3-VL-32B & 128,998 \\
\bottomrule
\end{tabular}
\caption{Median token usage per game in the canonical evaluation. Counts include the repeated perception--reasoning--action interaction and should be interpreted alongside provider-specific tokenization.}
\label{tab:token_usage}
\end{table}

As a scaffold-sensitivity check, we evaluated Claude Code with Opus 4.6 on all ten public games under the same 120-step budget. Its geometric-mean human-relative score was 12.46, compared with 5.91 for canonical Claude Opus 4.5, while median token use increased to 257,700 from 159,845. It shows that stronger agentic scaffolding can materially change the measured level and resource cost.

Human play was single-shot at the game level: every participant received ten distinct games, one 120-second episode per game, and no later session on the same game. Learning and retries could occur only within an episode. Model evaluation used provider APIs with reasoning and tool use enabled where supported. These details reduce ambiguity about experience and resource budgets while leaving open the broader question of how to equate human and machine compute.

\subsection{Human-in-the-loop effort and contamination considerations}\label{sec:effort_contamination}

Accepted games required approximately 30 minutes of human refinement on average after generation and automated checks. At that observed rate, a trained reviewer can process hundreds of candidates per month, and a small team can extend the suite to thousands over several months, although realized throughput will depend on quality thresholds and game complexity. The adaptations are independently generated JavaScript environments from game descriptions: source-title knowledge does not expose the exact layouts, dynamics, controls, or levels used for evaluation. The novel variants in Appendix \ref{sec:novel_variants} provide an additional route to reducing direct title-level contamination, but no benchmark can rule out all transfer from related games or general gameplay data.

\section{Model failure cases}\label{sec:failure_case}

In this section, we present some qualitative examples of gameplay by GPT-5.2. We observe the following failure patterns:

First, the model often attempt invalid actions and fail to learn from its mistakes. For example, in the game adapted from water sort puzzle, the model frequently tries to pour liquid from a tube to another tube with different colored-liquid, even if this is not allowed (top liquid color needs to match for a liquid to be transferred to another). While humans also make the same mistakes early in the game, they can quickly learn about how the game works and improve their strategy. GPT-5.2 often repeatedly make the same mistakes.

Second, models often struggle with visual reasoning. In the game inspired by Bubble Cash, we find that the model often do not shoot in the direction that would hit the ball of the same color, indicating that it has poor visual planning capability. In some cases, the model mistakenly claims that a tube is empty even if it's half-full.

Third, model struggle with physical reasoning and control. In our game adapted from angry birds, we find that the model often cannot estimate the right angle to shoot the birds. Very frequently, the model shoot in a horizontal angle and neglect the effect of gravity in the bird's trajectory. The model also frequently shoots in the opposite direction away from the targets.

These findings collectively suggest that models today still lack some of the basic cognitive capabilities to succeed in diverse set of digital human games.

\begin{figure}[ht!]
    \centering

    \label{fig:three_images}

    \begin{minipage}{0.35\textwidth}
        \centering
        \includegraphics[width=\linewidth]{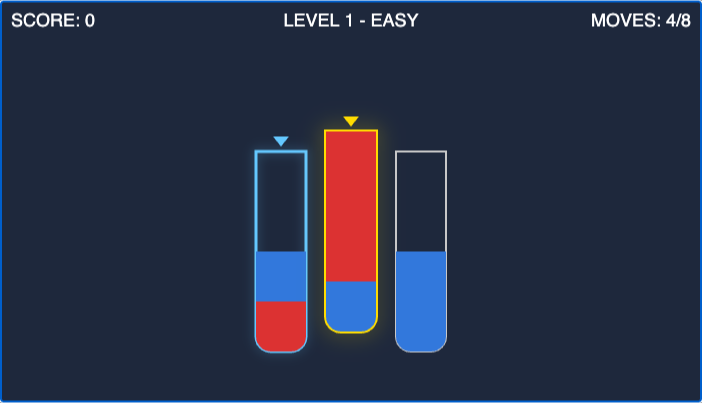}
        \label{fig:image1}
    \end{minipage}%
    \hfill%
    \begin{minipage}{0.24\textwidth}
        \centering
        \includegraphics[width=\linewidth]{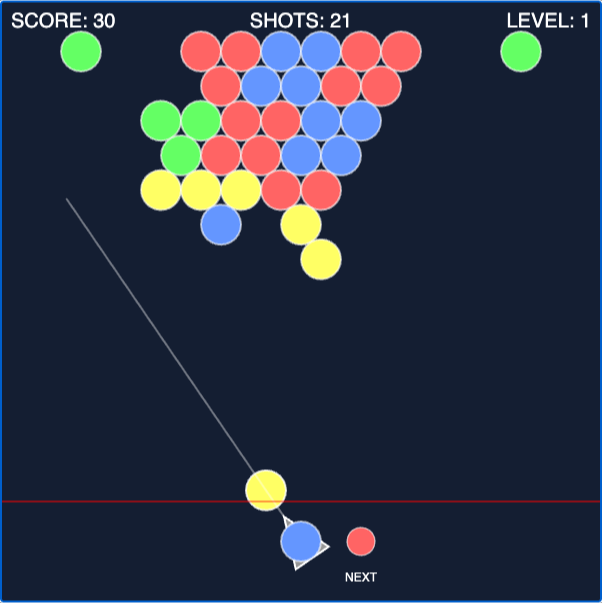}
        \label{fig:image2}
    \end{minipage}%
    \hfill%
    \begin{minipage}{0.35\textwidth}
        \centering
        \includegraphics[width=\linewidth]{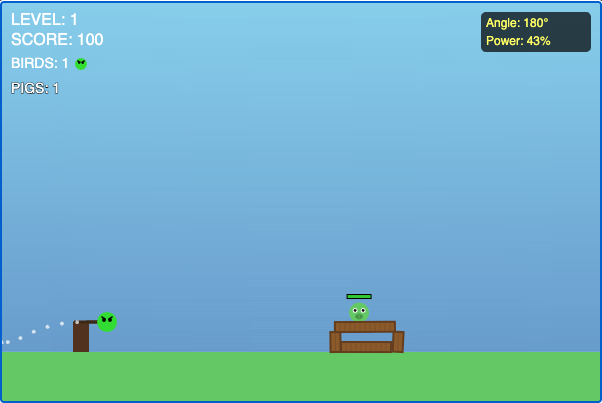}
        \label{fig:image3}
    \end{minipage}
    \caption{Example failure cases in \textbf{\textsc{AI GameStore}} games by GPT-5.2. The three games are adapted from Water Sort, Bubble Cash and Angry Birds. In game 1 (left), the model repeatably tries to pour from tube 2 to tube 1, which is an invalid action. In game 2 (middle), the model is not aiming the shots properly at the same color. In game 3 (right), the model shots in the opposite direction and with a horizontal angle, demonstrating poor understanding of physics.}
    \vspace{-1em}
\end{figure}

\newpage

\input{checklist.tex}

\end{document}

%% file: checklist.tex
\section*{NeurIPS Paper Checklist}

\begin{enumerate}

\item {\bf Claims}
    \item[] Question: Do the main claims made in the abstract and introduction accurately reflect the paper's contributions and scope?
    \item[] Answer: \answerYes{} %
    \item[] Justification: The paper abstract and introduction states the contribution of the paper.

\item {\bf Limitations}
    \item[] Question: Does the paper discuss the limitations of the work performed by the authors?
    \item[] Answer: \answerYes{} %
    \item[] Justification: The limitations are discussed in the Discussion section

\item {\bf Theory assumptions and proofs}
    \item[] Question: For each theoretical result, does the paper provide the full set of assumptions and a complete (and correct) proof?
    \item[] Answer: \answerNA{} %
    \item[] Justification: \answerNA{}

    \item {\bf Experimental result reproducibility}
    \item[] Question: Does the paper fully disclose all the information needed to reproduce the main experimental results of the paper to the extent that it affects the main claims and/or conclusions of the paper (regardless of whether the code and data are provided or not)?
    \item[] Answer: \answerYes{} %
    \item[] Justification: We release the game generation pipeline, the model evaluation harness, the human experiment code, the 10 public games, and the human study data at \url{https://github.com/lance-ying/aigamestore}, with setup instructions in each component's README. The public games can also be played and evaluated through \url{https://aigamestore.org}. Prompts, harness parameters, and scoring are described in Section~\ref{sec:experiments} and Appendix~\ref{sec:harness}.

\item {\bf Open access to data and code}
    \item[] Question: Does the paper provide open access to the data and code, with sufficient instructions to faithfully reproduce the main experimental results, as described in supplemental material?
    \item[] Answer: \answerYes{} %
    \item[] Justification: Code and data are openly available at \url{https://github.com/lance-ying/aigamestore} (MIT license), and the public games and leaderboard are at \url{https://aigamestore.org}. To limit benchmark contamination, 90 of the 100 games are held out as a private test set; access to the full benchmark is available upon request.

\item {\bf Experimental setting/details}
    \item[] Question: Does the paper specify all the training and test details (e.g., data splits, hyperparameters, how they were chosen, type of optimizer) necessary to understand the results?
    \item[] Answer: \answerYes{} %
    \item[] Justification: We described the experiment setting in detail, which includes prompts and parameters.

\item {\bf Experiment statistical significance}
    \item[] Question: Does the paper report error bars suitably and correctly defined or other appropriate information about the statistical significance of the experiments?
    \item[] Answer: \answerYes{} %
    \item[] Justification: Error bars are plotted in the results figures.

\item {\bf Experiments compute resources}
    \item[] Question: For each experiment, does the paper provide sufficient information on the computer resources (type of compute workers, memory, time of execution) needed to reproduce the experiments?
    \item[] Answer: \answerYes{} %
    \item[] Justification: We describe the model experiments in detail, including the number of inference steps, model specification etc. 
    
\item {\bf Code of ethics}
    \item[] Question: Does the research conducted in the paper conform, in every respect, with the NeurIPS Code of Ethics \url{https://neurips.cc/public/EthicsGuidelines}?
    \item[] Answer: \answerYes{} %
    \item[] Justification: This research comply with the NeurIPS code of ethics.

\item {\bf Broader impacts}
    \item[] Question: Does the paper discuss both potential positive societal impacts and negative societal impacts of the work performed?
    \item[] Answer: \answerNA{} %
    \item[] Justification: Our work aims to evaluate the capability of today's frontier models through a scalable game generation and evaluation platform. It currently does not produce any artifact that would have a direct societal impact.
    
\item {\bf Safeguards}
    \item[] Question: Does the paper describe safeguards that have been put in place for responsible release of data or models that have a high risk for misuse (e.g., pre-trained language models, image generators, or scraped datasets)?
    \item[] Answer: \answerNA{} %
    \item[] Justification: Our work does not produce any data or models that can be directly used for harmful applications.

\item {\bf Licenses for existing assets}
    \item[] Question: Are the creators or original owners of assets (e.g., code, data, models), used in the paper, properly credited and are the license and terms of use explicitly mentioned and properly respected?
    \item[] Answer: \answerYes{} %
    \item[] Justification: The convergent-validity analysis (Appendix~\ref{sec:convergent_validity}) uses existing public benchmarks, which we name and use under their original licenses and terms: PIQA, MMMU, MMToM-QA, BABILong, Blocksworld, ARC-AGI-1, and the publicly reported Artificial Analysis Intelligence Index. Source games from the Apple App Store and Steam are referenced by title only; the AI GameStore environments are independently generated JavaScript adaptations of public game descriptions, not copies of the original games.

\item {\bf New assets}
    \item[] Question: Are new assets introduced in the paper well documented and is the documentation provided alongside the assets?
    \item[] Answer: \answerYes{} %
    \item[] Justification: We introduce the AI GameStore benchmark: the game generation pipeline, evaluation harness, human experiment frontend, 10 public games, and human study data. These assets are documented in the repository READMEs and an MLCommons Croissant metadata file, and released at \url{https://github.com/lance-ying/aigamestore} under the MIT license (code) and CC BY 4.0 (data). The remaining 90 games are kept as a private test set.

\item {\bf Crowdsourcing and research with human subjects}
    \item[] Question: For crowdsourcing experiments and research with human subjects, does the paper include the full text of instructions given to participants and screenshots, if applicable, as well as details about compensation (if any)? 
    \item[] Answer: \answerYes{} %
    \item[] Justification: The human experiment frontend, the participant instructions, and the collected study data (with pseudonymized participant IDs) are available at \url{https://github.com/lance-ying/aigamestore}.

\item {\bf Institutional review board (IRB) approvals or equivalent for research with human subjects}
    \item[] Question: Does the paper describe potential risks incurred by study participants, whether such risks were disclosed to the subjects, and whether Institutional Review Board (IRB) approvals (or an equivalent approval/review based on the requirements of your country or institution) were obtained?
    \item[] Answer: \answerYes{} %
    \item[] Justification: The human experiment is approved by an IRB.

\item {\bf Declaration of LLM usage}
    \item[] Question: Does the paper describe the usage of LLMs if it is an important, original, or non-standard component of the core methods in this research? Note that if the LLM is used only for writing, editing, or formatting purposes and does \emph{not} impact the core methodology, scientific rigor, or originality of the research, declaration is not required.
    \item[] Answer: \answerYes{} %
    \item[] Justification: LLMs are used for game environment generation. The use of LLM and the pipeline is clearly described in the paper.

\end{enumerate}

%% file: neurips_2026.bbl
\begin{thebibliography}{10}

\bibitem{allen2024using}
Kelsey Allen, Franziska Br{\"a}ndle, Matthew Botvinick, Judith~E Fan, Samuel~J
  Gershman, Alison Gopnik, Thomas~L Griffiths, Joshua~K Hartshorne, Tobias~U
  Hauser, Mark~K Ho, et~al.
\newblock Using games to understand the mind.
\newblock {\em Nature Human Behaviour}, pages 1--9, 2024.

\bibitem{arcagi3_2026}
{{ARC Prize}}.
\newblock Arc-agi-3: The first interactive reasoning benchmark, 2026.
\newblock Accessed: February 12, 2026.

\bibitem{bubeck2023sparks}
S{\'e}bastien Bubeck, Varun Chandrasekaran, Ronen Eldan, Johannes Gehrke, Eric
  Horvitz, Ece Kamar, Peter Lee, Yin~Tat Lee, Yuanzhi Li, Scott Lundberg,
  et~al.
\newblock Sparks of artificial general intelligence: Early experiments with
  gpt-4.
\newblock {\em arXiv preprint arXiv:2303.12712}, 2023.

\bibitem{mlayoutstutorial2024}
John Burden, Lucy Cheke, Jos{\'e} Hern{\'a}ndez-Orallo, Marija Te{\v{s}}i{\'c},
  and Konstantinos Voudouris.
\newblock Measurement layouts for capability-oriented {AI} evaluation.
\newblock AAAI 2024 Tutorial, 2024.
\newblock Presented at the 38th AAAI Conference on Artificial Intelligence
  (AAAI-24).

\bibitem{burghardt2024animal}
Gordon~M Burghardt, Sergio~M Pellis, Jeffrey~C Schank, Paul~E Smaldino,
  Louk~JMJ Vanderschuren, and Elisabetta Palagi.
\newblock Animal play and evolution: Seven timely research issues about
  enigmatic phenomena.
\newblock {\em Neuroscience \& Biobehavioral Reviews}, 160:105617, 2024.

\bibitem{Campbell2002Deep}
Murray Campbell, A~Joseph Hoane~Jr, and Feng-hsiung Hsu.
\newblock Deep blue.
\newblock {\em Artificial intelligence}, 134(1-2):57--83, 2002.

\bibitem{chater2018mind}
Nick Chater, Teppo Felin, David~C Funder, Gerd Gigerenzer, Jan~J Koenderink,
  Joachim~I Krueger, Denis Noble, Samuel~A Nordli, Mike Oaksford, Barry
  Schwartz, et~al.
\newblock Mind, rationality, and cognition: An interdisciplinary debate.
\newblock {\em Psychonomic Bulletin \& Review}, 25(2):793--826, 2018.

\bibitem{chen2026does}
Eddy~Keming Chen, Mikhail Belkin, Leon Bergen, and David Danks.
\newblock Does {AI} already have human-level intelligence? the evidence is
  clear.
\newblock {\em Nature}, 650(8100):36--40, 2026.

\bibitem{chu2020play}
Junyi Chu and Laura~E Schulz.
\newblock Play, curiosity, and cognition.
\newblock {\em Annual Review of Developmental Psychology}, 2(1):317--343, 2020.

\bibitem{chu2024praise}
Junyi Chu, Joshua~B Tenenbaum, and Laura~E Schulz.
\newblock In praise of folly: flexible goals and human cognition.
\newblock {\em Trends in Cognitive Sciences}, 28(7):628--642, 2024.

\bibitem{cleveland1907psychology}
Alfred~A Cleveland.
\newblock The psychology of chess and of learning to play it.
\newblock {\em The American Journal of Psychology}, 18(3):269--308, 1907.

\bibitem{cobbe2020leveraging}
Karl Cobbe, Chris Hesse, Jacob Hilton, and John Schulman.
\newblock Leveraging procedural generation to benchmark reinforcement learning.
\newblock In {\em International conference on machine learning}, pages
  2048--2056. PMLR, 2020.

\bibitem{Cobbe2021Training}
Karl Cobbe, Vineet Kosaraju, Mohammad Bavarian, Mark Chen, Heewoo Jun, Lukasz
  Kaiser, Matthias Plappert, Jerry Tworek, Jacob Hilton, Reiichiro Nakano,
  et~al.
\newblock Training verifiers to solve math word problems.
\newblock {\em arXiv preprint arXiv:2110.14168}, 2021.

\bibitem{collins2026expert}
Katherine~M Collins and Joshua~B Tenenbaum.
\newblock Expert-level test is a head-scratcher for {AI}, 2026.

\bibitem{collins2025gamecreation}
Katherine~M Collins, Graham Todd, Lionel Wong, Cedeago Zhang, Julius Togelius,
  Adrian Weller, Junyi Chu, Thomas Griffiths, and Josh Tenenbaum.
\newblock Generation and evaluation in the human invention process through the
  lens of game design.
\newblock In {\em Proceedings of the Annual Meeting of the Cognitive Science
  Society}, volume~47, 2025.

\bibitem{collins2025people}
Katherine~M Collins, Cedegao~E Zhang, Lionel Wong, Mauricio~Barba da~Costa,
  Graham Todd, Adrian Weller, Samuel~J Cheyette, Thomas~L Griffiths, and
  Joshua~B Tenenbaum.
\newblock People use fast, flat goal-directed simulation to reason about novel
  problems.
\newblock {\em arXiv preprint arXiv:2510.11503}, 2025.

\bibitem{feng2007playing}
Jing Feng, Ian Spence, and Jay Pratt.
\newblock Playing an action video game reduces gender differences in spatial
  cognition.
\newblock {\em Psychological science}, 18(10):850--855, 2007.

\bibitem{genesereth2005general}
Michael Genesereth, Nathaniel Love, and Barney Pell.
\newblock General game playing: Overview of the aaai competition.
\newblock {\em AI magazine}, 26(2):62--62, 2005.

\bibitem{gobet2004moves}
Fernand Gobet, Jean Retschitzki, and Alex de~Voogt.
\newblock {\em Moves in mind: The psychology of board games}.
\newblock Psychology Press, 2004.

\bibitem{google2026gamearena}
Google.
\newblock Advancing {AI} benchmarking with game arena.
\newblock Google DeepMind Blog, February 2026.
\newblock Accessed: February 12, 2026.

\bibitem{hafner2021benchmarking}
Danijar Hafner.
\newblock Benchmarking the spectrum of agent capabilities.
\newblock {\em arXiv preprint arXiv:2109.06780}, 2021.

\bibitem{hendrycks2025definition}
Dan Hendrycks, Dawn Song, Christian Szegedy, Honglak Lee, Yarin Gal, Erik
  Brynjolfsson, Sharon Li, Andy Zou, Lionel Levine, Bo~Han, et~al.
\newblock A definition of agi.
\newblock {\em arXiv preprint arXiv:2510.18212}, 2025.

\bibitem{hernandez2017new}
Jos{\'e} Hern{\'a}ndez-Orallo, Marco Baroni, Jordi Bieger, Nader Chmait,
  David~L Dowe, Katja Hofmann, Fernando Mart{\'\i}nez-Plumed, Claes
  Stranneg{\aa}rd, and Kristinn~R Th{\'o}risson.
\newblock A new {AI} evaluation cosmos: Ready to play the game?
\newblock {\em AI Magazine}, 38(3):66--69, 2017.

\bibitem{jimenez2023swe}
Carlos~E Jimenez, John Yang, Alexander Wettig, Shunyu Yao, Kexin Pei, Ofir
  Press, and Karthik Narasimhan.
\newblock Swe-bench: Can language models resolve real-world github issues?
\newblock {\em arXiv preprint arXiv:2310.06770}, 2023.

\bibitem{kanervisto2025world}
Anssi Kanervisto, Dave Bignell, Linda~Yilin Wen, Martin Grayson, Raluca
  Georgescu, Sergio Valcarcel~Macua, Shan~Zheng Tan, Tabish Rashid, Tim Pearce,
  Yuhan Cao, et~al.
\newblock World and human action models towards gameplay ideation.
\newblock {\em Nature}, 638(8051):656--663, 2025.

\bibitem{legg2007universal}
Shane Legg and Marcus Hutter.
\newblock Universal intelligence: A definition of machine intelligence.
\newblock {\em Minds and machines}, 17(4):391--444, 2007.

\bibitem{lehrach2025code}
Wolfgang Lehrach, Daniel Hennes, Miguel Lazaro-Gredilla, Xinghua Lou, Carter
  Wendelken, Zun Li, Antoine Dedieu, Jordi Grau-Moya, Marc Lanctot, Atil Iscen,
  et~al.
\newblock Code world models for general game playing.
\newblock {\em arXiv preprint arXiv:2510.04542}, 2025.

\bibitem{leibo2021scalable}
Joel~Z Leibo, Edgar~A Due{\~n}ez-Guzman, Alexander Vezhnevets, John~P Agapiou,
  Peter Sunehag, Raphael Koster, Jayd Matyas, Charlie Beattie, Igor Mordatch,
  and Thore Graepel.
\newblock Scalable evaluation of multi-agent reinforcement learning with
  melting pot.
\newblock In {\em International conference on machine learning}, pages
  6187--6199. PMLR, 2021.

\bibitem{li2025gvgai}
Yuchen Li, Cong Lin, Muhammad~Umair Nasir, Philip Bontrager, Jialin Liu, and
  Julian Togelius.
\newblock Gvgai-llm: Evaluating large language model agents with infinite
  games.
\newblock {\em arXiv preprint arXiv:2508.08501}, 2025.

\bibitem{magne2026nitrogen}
Lo{\"\i}c Magne, Anas Awadalla, Guanzhi Wang, Yinzhen Xu, Joshua Belofsky,
  Fengyuan Hu, Joohwan Kim, Ludwig Schmidt, Georgia Gkioxari, Jan Kautz, et~al.
\newblock Nitrogen: An open foundation model for generalist gaming agents.
\newblock {\em arXiv preprint arXiv:2601.02427}, 2026.

\bibitem{mitchell2024debates}
Melanie Mitchell.
\newblock Debates on the nature of artificial general intelligence, 2024.

\bibitem{nasir2024word2world}
Muhammad~U Nasir, Steven James, and Julian Togelius.
\newblock Word2world: Generating stories and worlds through large language
  models.
\newblock {\em arXiv preprint arXiv:2405.06686}, 2024.

\bibitem{newell1965example}
A~Newell and HA~Simon.
\newblock An example of human chess play in the light of chess playing
  programs.
\newblock {\em Progress in Biocybernetics}, 2:19--75, 1965.

\bibitem{newell1972human}
Allen Newell, Herbert~Alexander Simon, et~al.
\newblock {\em Human problem solving}, volume 104.
\newblock Prentice-hall Englewood Cliffs, NJ, 1972.

\bibitem{OpenAIUniverse2016}
{OpenAI}.
\newblock Universe.
\newblock \url{https://openai.com/index/universe/}, December 2016.
\newblock Archived content; Accessed: December 3, 2025.

\bibitem{paglieri2024balrog}
Davide Paglieri, Bart{\l}omiej Cupia{\l}, Samuel Coward, Ulyana Piterbarg,
  Maciej Wolczyk, Akbir Khan, Eduardo Pignatelli, {\L}ukasz Kuci{\'n}ski,
  Lerrel Pinto, Rob Fergus, et~al.
\newblock Balrog: Benchmarking agentic llm and vlm reasoning on games.
\newblock {\em arXiv preprint arXiv:2411.13543}, 2024.

\bibitem{perez2019general}
Diego Perez-Liebana, Jialin Liu, Ahmed Khalifa, Raluca~D Gaina, Julian
  Togelius, and Simon~M Lucas.
\newblock General video game ai: A multitrack framework for evaluating agents,
  games, and content generation algorithms.
\newblock {\em IEEE Transactions on Games}, 11(3):195--214, 2019.

\bibitem{phan2025humanity}
Long Phan, Alice Gatti, Ziwen Han, Nathaniel Li, Josephina Hu, Hugh Zhang, Chen
  Bo~Calvin Zhang, Mohamed Shaaban, John Ling, Sean Shi, et~al.
\newblock Humanity's last exam.
\newblock {\em arXiv preprint arXiv:2501.14249}, 2025.

\bibitem{schaul2011measuring}
Tom Schaul, Julian Togelius, and J{\"u}rgen Schmidhuber.
\newblock Measuring intelligence through games.
\newblock {\em arXiv preprint arXiv:1109.1314}, 2011.

\bibitem{Silver2016Mastering}
David Silver, Aja Huang, Chris~J Maddison, Arthur Guez, Laurent Sifre, George
  Van Den~Driessche, Julian Schrittwieser, Ioannis Antonoglou, Veda
  Panneershelvam, Marc Lanctot, et~al.
\newblock Mastering the game of go with deep neural networks and tree search.
\newblock {\em nature}, 529(7587):484--489, 2016.

\bibitem{smith1982does}
Peter~K Smith.
\newblock Does play matter? functional and evolutionary aspects of animal and
  human play.
\newblock {\em Behavioral and brain sciences}, 5(1):139--155, 1982.

\bibitem{spence2010video}
Ian Spence and Jing Feng.
\newblock Video games and spatial cognition.
\newblock {\em Review of general psychology}, 14(2):92--104, 2010.

\bibitem{srivastava2023beyond}
Aarohi Srivastava, Abhinav Rastogi, Abhishek Rao, Abu Awal~Md Shoeb, Abubakar
  Abid, Adam Fisch, Adam~R Brown, Adam Santoro, Aditya Gupta, Adri{\`a}
  Garriga-Alonso, et~al.
\newblock Beyond the imitation game: Quantifying and extrapolating the
  capabilities of language models.
\newblock {\em Transactions on machine learning research}, 2023.

\bibitem{todd2023level}
Graham Todd, Sam Earle, Muhammad~Umair Nasir, Michael~Cerny Green, and Julian
  Togelius.
\newblock Level generation through large language models.
\newblock In {\em Proceedings of the 18th International Conference on the
  Foundations of Digital Games}, pages 1--8, 2023.

\bibitem{todd2024gavel}
Graham Todd, Alexander~G Padula, Matthew Stephenson, {\'E}ric Piette, Dennis~J
  Soemers, and Julian Togelius.
\newblock Gavel: Generating games via evolution and language models.
\newblock {\em Advances in Neural Information Processing Systems},
  37:110723--110745, 2024.

\bibitem{tsividis2021human}
Pedro~A Tsividis, Joao Loula, Jake Burga, Nathan Foss, Andres Campero, Thomas
  Pouncy, Samuel~J Gershman, and Joshua~B Tenenbaum.
\newblock Human-level reinforcement learning through theory-based modeling,
  exploration, and planning.
\newblock {\em arXiv preprint arXiv:2107.12544}, 2021.

\bibitem{turing1950computing}
Alan~M. Turing.
\newblock Computing machinery and intelligence.
\newblock {\em Mind}, LIX(236):433--460, 1950.

\bibitem{verma2025measuring}
Vivek Verma, David Huang, William Chen, Dan Klein, and Nicholas Tomlin.
\newblock Measuring general intelligence with generated games.
\newblock {\em arXiv preprint arXiv:2505.07215}, 2025.

\bibitem{wang2018glue}
Alex Wang, Amanpreet Singh, Julian Michael, Felix Hill, Omer Levy, and Samuel
  Bowman.
\newblock Glue: A multi-task benchmark and analysis platform for natural
  language understanding.
\newblock In {\em Proceedings of the 2018 EMNLP workshop BlackboxNLP: Analyzing
  and interpreting neural networks for NLP}, pages 353--355, 2018.

\bibitem{wang2025game}
Zihao Wang, Xujing Li, Yining Ye, Junjie Fang, Haoming Wang, Longxiang Liu,
  Shihao Liang, Junting Lu, Zhiyong Wu, Jiazhan Feng, et~al.
\newblock Game-tars: Pretrained foundation models for scalable generalist
  multimodal game agents.
\newblock {\em arXiv preprint arXiv:2510.23691}, 2025.

\bibitem{warrier2025benchmarking}
Archana Warrier, Dat Nguyen, Michelangelo Naim, Moksh Jain, Yichao Liang, Karen
  Schroeder, Cambridge Yang, Joshua~B Tenenbaum, Sebastian Vollmer, Kevin
  Ellis, et~al.
\newblock Benchmarking world-model learning.
\newblock {\em arXiv preprint arXiv:2510.19788}, 2025.

\bibitem{xing2024understanding}
Mingzhe Xing, Rongkai Zhang, Hui Xue, Qi~Chen, Fan Yang, and Zhen Xiao.
\newblock Understanding the weakness of large language model agents within a
  complex android environment.
\newblock In {\em Proceedings of the 30th ACM SIGKDD Conference on Knowledge
  Discovery and Data Mining}, pages 6061--6072, 2024.

\bibitem{ying2025assessing}
Lance Ying, Katherine~M Collins, Prafull Sharma, Cedric Colas, Kaiya~Ivy Zhao,
  Adrian Weller, Zenna Tavares, Phillip Isola, Samuel~J Gershman, Jacob~D
  Andreas, et~al.
\newblock Assessing adaptive world models in machines with novel games.
\newblock {\em arXiv preprint arXiv:2507.12821}, 2025.

\bibitem{zhang2025videogamebench}
Alex~L Zhang, Thomas~L Griffiths, Karthik~R Narasimhan, and Ofir Press.
\newblock Videogamebench: Can vision-language models complete popular video
  games?
\newblock {\em arXiv preprint arXiv:2505.18134}, 2025.

\bibitem{zhou2025general}
Lexin Zhou, Lorenzo Pacchiardi, Fernando Mart{\'\i}nez-Plumed, Katherine~M
  Collins, Yael Moros-Daval, Seraphina Zhang, Qinlin Zhao, Yitian Huang, Luning
  Sun, Jonathan~E Prunty, et~al.
\newblock General scales unlock {AI} evaluation with explanatory and predictive
  power.
\newblock {\em arXiv preprint arXiv:2503.06378}, 2025.

\end{thebibliography}
